\documentclass{article}

\PassOptionsToPackage{numbers, compress}{natbib}
 \usepackage[preprint]{neurips_2026}

\usepackage[utf8]{inputenc} % allow utf-8 input
\usepackage[T1]{fontenc}    % use 8-bit T1 fonts
\usepackage{hyperref}       % hyperlinks
\usepackage{xurl}           % URL typesetting; allows line breaks anywhere so long URLs wrap
\usepackage{booktabs}       % professional-quality tables
\usepackage{amsmath}        % math packages
\usepackage{amsfonts}       % blackboard math symbols
\usepackage{nicefrac}       % compact symbols for 1/2, etc.
\usepackage{microtype}      % microtypography
\usepackage{xcolor}         % colors
\usepackage{colortbl}       % colored table rules (\arrayrulecolor)
\usepackage{graphicx}       % figures
\usepackage{subcaption}     % subfigures
\usepackage{tikz}           % inline diagrams
\usetikzlibrary{arrows.meta, decorations.pathreplacing, positioning, calc}

\title{Are Sparse Autoencoder Benchmarks Reliable?}

\author{%
  David Chanin \\
  Decode Research, MATS, UCL \\
  \texttt{david.chanin.22@ucl.ac.uk} \\
}

\begin{document}

\maketitle

\begin{abstract}
Sparse autoencoders (SAEs) are a core interpretability tool for large language models, and progress on SAE architectures depends on benchmarks that reliably distinguish better SAEs from worse ones. We audit the SAE quality metrics in SAEBench, the de-facto standard SAE evaluation suite, through three complementary lenses: reseed noise on a fixed SAE, ground-truth correlation on synthetic SAEs, and discriminability across training trajectories. We find that two of these metrics, Targeted Probe Perturbation (TPP) and Spurious Correlation Removal (SCR), fail multiple lenses at their canonical settings and should not be used to evaluate SAEs. The other metrics show higher reseed noise and lower discriminability than the field assumes. The sae-probes variant of $k$-sparse probing is the most reliable metric we tested, but even sae-probes struggles to separate variants of the same SAE architecture. Our results show the field needs better SAE benchmarks.
\end{abstract}

\section{Introduction}
\label{sec:intro}

Sparse autoencoders (SAEs) decompose a neural network's activation vector into a sparse sum of feature directions and are a core mechanistic interpretability technique for large language models (LLMs)~\citep{huben2024sparse,bricken2023towards}. However, recent work has called into question the quality of SAEs compared to supervised baselines~\citep{kantamneni2025sparse,wu2025axbench}, highlighting the need for better SAE architectures.

Benchmarking SAEs is harder than most ML evaluation: the ``true features'' of an LLM are unknown, so we cannot directly tell how well an SAE has learned these ``true features''. Instead, the field relies on a number of proxy metrics collated in the SAEBench library~\citep{karvonen2025saebench}. Architectural progress comes from small comparisons (a different sparsity penalty, a change to the activation function, an improved auxiliary loss), each staking its case on these proxy metrics, with subsequent work building on whichever variant won. When the proxies rank SAEs reliably, this loop compounds into real progress. When they do not, effort flows into changes that look better on a noisy proxy but are not, and real improvements go undetected. Auditing the proxies is therefore a prerequisite to trusting the architectures they rank.

While directly auditing proxy metrics is difficult due to not knowing the ground truth ``true features'' of an LLM, we can still define properties that a good proxy metric should satisfy. These include: 

\begin{enumerate}
  \item \textbf{Tracks ground-truth.} The score should be monotone in the SAE's ground-truth quality if ground truth is available (e.g. on synthetic models).
  \item \textbf{Increase throughout training.} A metric whose score systematically declines through training is responding to something other than quality.
  \item \textbf{Low reseed noise.} The same SAE evaluated on the same data should produce the same score regardless of any random seed used.
  \item \textbf{Discriminative.} The metric should distinguish runs by an amount substantially larger than its own per-checkpoint jitter.
  \item \textbf{Internally consistent.} The metric should rank SAEs the same way regardless of nominally-interchangeable hyperparameters.
\end{enumerate}

In this work, we audit six SAEBench evaluations that target overall SAE quality: sparse probing~\citep{gurnee2023finding}, the sae-probes variant of sparse probing~\citep{kantamneni2025sparse}, targeted probe perturbation (TPP)~\citep{marks2025sparse}, spurious correlation removal (SCR)~\citep{marks2025sparse}, automated interpretation~\citep{bills2023language}, and RAVEL~\citep{huang2024ravel}. The five desiderata above cannot all be tested on a single SAE panel: low reseed noise needs the same SAE evaluated many times, ground-truth tracking needs SAEs whose ground-truth quality is known, and discriminability needs SAEs that vary in quality. We therefore approach the audit through three complementary lenses, each on a different SAE panel and each isolating a different failure mode:

\textbf{Reseed noise on a real LLM SAE} (\S\ref{sec:noise}, desideratum~3). We run each SAEBench evaluation five times with different random seeds on a fixed canonical Gemma Scope SAE~\citep{lieberumGemmaScopeOpen2024} (replicated on three other canonical SAEs across model families in Appendix~\ref{app:per_sae_noise}). Because the SAE is identical across runs, any score spread is pure metric noise; the resulting per-metric CV spans nearly two orders of magnitude and yields a minimum-reliable-difference threshold for single-seed comparisons.

\textbf{Validity on synthetic SAEs} (\S\ref{sec:gt}, desideratum~1). Using SynthSAEBench-16k~\citep{chanin2026synthsaebench}, a synthetic model whose activations are sparse linear combinations of a known 16k-feature dictionary, we train a panel of SAEs with computable ground-truth quality and check whether each proxy metric's score correlates with that ground truth. This is the only setting in which we can adjudicate desideratum~1 directly; autointerp and RAVEL cannot be tested in this lens because both require natural-language concepts the synthetic dictionary lacks.

\textbf{Discriminability across training} (\S\ref{sec:snr}, desiderata~2, 4, and 5). We train two panels of SAEs and snapshot them throughout training: a four-SAE \emph{cross-architecture} panel with deliberately large differences (BatchTopK vs Matryoshka, $k\!\in\!\{50, 100\}$), and a four-SAE \emph{Matryoshka} panel that varies a single hyperparameter $n$ across three independent training seeds per variant. The cross-architecture panel asks whether the metric can tell apart very different SAEs; the Matryoshka panel asks the harder, more practically relevant question of whether it can tell apart the small hyperparameter variants practitioners typically compare. The training-trajectory axis lets us check desideratum~2 directly, and the three Matryoshka seeds let us decompose total score variance into between-SAE signal versus training-seed and snapshot noise.

% A metric must pass all three lenses to be trustworthy. We observe metrics that fail each in isolation, and TPP and SCR fail multiple lenses simultaneously.

Across the three lenses, TPP and SCR fail multiple desiderata at canonical settings: TPP scores \emph{worse} the more an SAE is trained, and SCR becomes negatively correlated with ground-truth at large top-$N$. Every other metric is noisier and less discriminative than the field assumes. The sae-probes variant of sparse probing is the most reliable metric we audit; its two main design choices, a much larger dataset suite (113 vs 5) and cross-validated probe regularization, suggest a concrete path forward for the rest of the suite.

Our code is available via Github\footnote{Code: \url{https://github.com/decoderesearch/saebench-reliability-audit}}, and SAE panels are available on Huggingface\footnote{SAE panels: \url{https://huggingface.co/decoderesearch/sae-snapshot-panels}}.

\section{Background}
\label{sec:background}

\paragraph{Sparse autoencoder} A sparse autoencoder maps an activation vector $x \in \mathbb{R}^d$ to a sparse latent vector $z = \sigma(W_{\text{enc}} x + b_{\text{enc}}) \in \mathbb{R}^{d_{\text{sae}}}$ with $d_{\text{sae}} \geq d$, and reconstructs $\hat{x} = W_{\text{dec}} z + b_{\text{dec}}$. SAEs are trained to minimize reconstruction error subject to a sparsity penalty on $z$, with the goal of recovering interpretable feature directions in the underlying model's activation space~\citep{bricken2023towards}. By ``SAE quality'' we mean the degree to which the trained latent directions align with semantically meaningful features.

\subsection{SAEBench metrics}

SAEBench~\citep{karvonen2025saebench} measures quality through six proxy metrics\footnote{We do not audit the unlearning metric as it requires chat models while the other metrics use base models. We do not audit absorption and meta structure metrics as these are meant to diagnose a specific failure and are not general quality metrics.}. Unless otherwise specified, we leave these at SAEBench default settings. Detailed descriptions of each metric are in Appendix~\ref{app:metric_details}.

\paragraph{Sparse probing} Trains a $k$-sparse logistic regression probe~\citep{gurnee2023finding} on the SAE latents to classify natural-language attributes such as language or profession; higher accuracy at small $k$ means concepts align with small numbers of latents. This benchmark uses 5 probing datasets.

\paragraph{Sae-probes sparse-probing} A new metric added to SAEBench after publication, this is a variation of sparse-probing based on \citet{kantamneni2025sparse}. This variant greatly increases the number of datasets used to 113, and uses L1-regularized logistic regression with cross-validation.

\paragraph{Spurious Correlation Removal (SCR)} Based on the SHIFT method from \citet{marks2025sparse}, trains intentionally biased probes on co-occurring concepts (e.g. all doctors are male) and measures how much that bias is removed when the SAE latents most associated with the spurious attribute are ablated. This benchmark uses 2 datasets.

\paragraph{Targeted Probe Perturbation (TPP)} Also based on \citet{marks2025sparse}, TPP trains a probe on each class within the same category (e.g. different profession), identifies the SAE latents whose ablation most damages the probe, and measures the gap in damage between the intended class probe and the unrelated probes when those top-$N$ latents are zeroed. This benchmark uses 2 datasets.

\paragraph{Autointerp} Based on \citet{paulo2024automatically}, Autointerp first randomly selects $N$ latents from the SAE, and uses an LLM to generate labels for those latents. Then, based on that description, a second LLM predicts whether or not that latent will fire on held-out explanations.

\paragraph{RAVEL} Based on \citet{huang2024ravel}, RAVEL evaluates disentanglement, causal effect, and isolation of latents corresponding to known relational properties. The metric tries to use SAE latents to causally change predicted attributes for target entities without affecting related entities.

\section{Reseed noise on a real LLM SAE}
\label{sec:noise}

To measure reseed noise we ran each SAEBench evaluation five times on a canonical Gemma Scope JumpReLU SAE (residual stream layer 12 of Gemma-2-2b, width 65k)~\citep{lieberumGemmaScopeOpen2024,gemmateam2024gemma2improvingopen}. Each run uses a different random seed and disables activation caching so that each seed is truly independent.

\textbf{Per-metric reseed noise.} Table~\ref{tab:noise} summarizes coefficient of variation (CV) on the canonical SAE for a curated subset of metrics. The noise structure is consistent across SAE families and the full per-$k$ and per-threshold breakdown (including the fixed-latent autointerp variant) is in Appendix~\ref{app:per_sae_noise}.

\begin{table}[h]
\caption{Reseed noise on the canonical 65k Gemma Scope SAE (Gemma-2-2b layer 12), five reseeds. The last column is the minimum-reliable-difference threshold $|\Delta|^*$ for a 95\% two-tailed comparison of two single-seed scores. Curated subset; the full per-$k$ and per-threshold table is in Appendix~\ref{app:per_sae_noise}.}
\label{tab:noise}
\centering
\small
\begin{tabular}{lrrrr}
\toprule
Benchmark & Mean & Std & CV & Min reliable $|\Delta|^*$ \\
\midrule
sae-probes ($k=5$, accuracy)            & 0.800 & 0.002 & 0.2\% & 0.008 \\
sae-probes ($k=1$, accuracy)            & 0.763 & 0.001 & 0.2\% & 0.004 \\
RAVEL disentangle                        & 0.706 & 0.001 & 0.2\% & 0.004 \\
sparse\_probing ($k=5$, accuracy)       & 0.852 & 0.003 & 0.3\% & 0.012 \\
autointerp                              & 0.839 & 0.004 & 0.5\% & 0.016 \\
sparse\_probing ($k=1$, accuracy)       & 0.722 & 0.009 & 1.2\% & 0.035 \\
RAVEL cause                              & 0.676 & 0.019 & 2.8\% & 0.075 \\
TPP top-500                              & 0.345 & 0.010 & 2.8\% & 0.039 \\
SCR top-10                               & 0.174 & 0.008 & 4.4\% & 0.031 \\
SCR top-500                              & 0.351 & 0.022 & 6.2\% & 0.086 \\
TPP top-50                               & 0.094 & 0.021 & 23\% & 0.083 \\
\bottomrule
\end{tabular}

\end{table}

The spread in noise CV is wide. sae-probes ($k=5$), sparse probing ($k=5$), and RAVEL disentangle are at CV $\leq 0.3\%$, low enough that single-seed comparisons resolve score deltas at or above $\sim 0.01$. SCR has CV 3--6\%; TPP has 16--39\% CV at small top-$N$ where the score itself is close to zero, so single-seed comparisons read mostly noise. The fifth column $|\Delta|^*$ reports the minimum-reliable difference for a 95\% two-tailed comparison of two single-seed scores, computed as $t_{0.025,\,4}\cdot s\sqrt{2}\approx 3.93\,s$ where $s$ is the column-3 std (derivation, multi-seed shrinkage, and cross-SAE caveats in Appendix~\ref{app:threshold_derivation}). Sparse probing, sae-probes, autointerp, and RAVEL disentangle have thresholds well below $0.04$ and are reliable single-seed for almost any plausible comparison; SCR (especially at large top-$N$) and TPP at moderate top-$N$ require multi-seed averaging.

\section{Discriminability across training trajectories}
\label{sec:snr}

To check whether SAEBench metrics can actually distinguish SAEs, we train two panels with different settings, snapshot them throughout training, and run all SAEBench metrics on each snapshot. The training-trajectory axis additionally surfaces clear failures like metrics that decline as the SAE trains.

\paragraph{Cross-architecture SAE set} We train 4 SAEs (BatchTopK $k\!\in\!\{50, 100\}$, Matryoshka $k\!\in\!\{50, 100\}$) for 1.5B tokens each on Gemma-2-2b layer 12, with LR decay over the final $1/5$, taking 28 checkpoints per SAE. These SAEs are deliberately very different; if SAEBench cannot differentiate them, the metrics have a serious problem.

\paragraph{Matryoshka SAE set} We train 4 Matryoshka SAEs~\citep{bussmann2025learning} varying in the number of Matryoshka levels per SAE, identified as $n \in \{1, 2, 3, 4\}$, and train each variant from 3 independent random seeds (12 SAE training runs total) so we can separate variant signal from training-seed noise. We train for 300M tokens on Gemma-2-2b layer 12 with LR decay over the final $1/5$, and take 10 post-warmup snapshots per (variant, seed).

\subsection{Results}

\begin{figure}[t]
\centering
\includegraphics[width=1\linewidth]{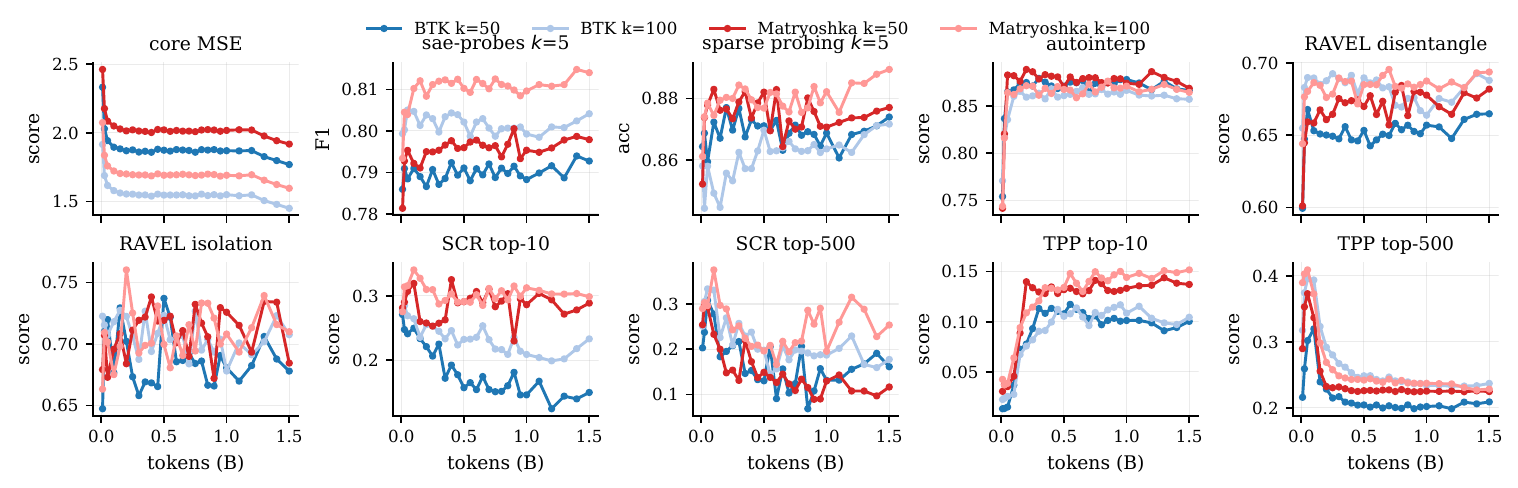}
\caption{Trajectories of ten representative metrics across the 1.5B-token training run in the cross-architecture SAE set. SCR and TPP top-500 both show that training the SAE makes the metric worse, a clear red-flag for metric quality. Metrics demonstrate varying levels of discriminability, but no metric approaches the clear separation and stability of MSE.}
\label{fig:training_traces}
\end{figure}

\begin{figure}[t]
\centering
\includegraphics[width=1\linewidth]{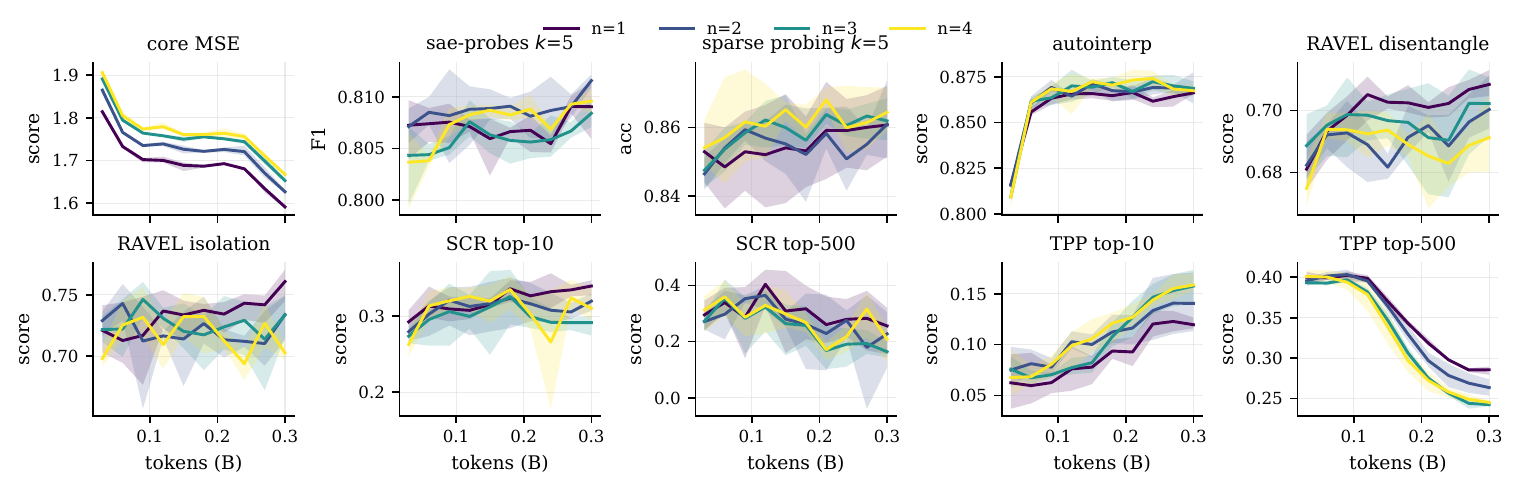}
\caption{Trajectories of the same ten headline metrics as Figure~\ref{fig:training_traces}, on the four-SAE Matryoshka set ($n\in\{1,\ldots,4\}$ inner widths sampled per training step, 300M tokens, 3 training seeds per variant). Solid lines are seed means; shaded bands are min/max envelopes across the 3 seeds. top-500 SCR and TPP show problematic declines over training. No non-core quality metric clearly differentiates these SAEs once seed envelopes are accounted for; multi-seed analysis is in Appendix~\ref{app:multi_seed}.}
\label{fig:sampled_mat_headline}
\end{figure}

We plot a representative subset of proxy metrics vs training tokens for the cross-architecture SAE set in Figure~\ref{fig:training_traces}, and for the Matryoshka SAE set in Figure~\ref{fig:sampled_mat_headline}. We include the core reconstruction Mean Squared Error (MSE) as a reference. Full results are shown in Appendix~\ref{app:full_traces}. Further analysis of metric training jitter and discriminability is in Appendix~\ref{app:snr_full}.

\paragraph{SCR and TPP decrease during training} The most clearly problematic metrics are again SCR and TPP. At top-500, both SCR and TPP actually \emph{decline} during training, implying the clearly incorrect conclusion that an untrained SAE is better than a trained SAE. The decline is not limited to high top-$N$: SCR also declines through training at the canonical top-$N=10$ on 2 of 4 SAEs in the cross-architecture panel (BatchTopK $k\!=\!50$ and $k\!=\!100$, slopes $-0.065$ and $-0.033$ score units per billion training tokens; Appendix~\ref{app:trajectories}, Table~\ref{tab:slopes}), so a practitioner running SCR at canonical settings on a standard architecture can observe the metric say their SAE is getting worse as training progresses. In addition, we do not see SCR and TPP ranking SAEs consistently between low and high values of top-$N$, making it hard to trust these metrics.

\paragraph{No metric reaches the discriminability of MSE} The inter-checkpoint jitter for all metrics is still much higher than it is for MSE, indicating there is still large room for improvement in metrics. For the cross-architecture SAE set, sae-probes does a good job of differentiating the different SAEs throughout training. For the Matryoshka SAE set, the four core reconstruction metrics, sae-probes ($k\!\in\!\{1, 5\}$), and RAVEL disentangle carry a real but small inter-variant signal (next paragraph); every other non-core metric, including all canonical SCR and TPP thresholds, is dominated by training-seed and snapshot noise rather than by the SAE variant. SAEBench is therefore not yet discerning enough to clearly guide most SAE hyperparameter decisions.

\paragraph{Variance decomposition on the Matryoshka panel} The three training seeds per variant let us split each metric's score variance into three sources: the SAE variant we picked, training-seed randomness within a variant, and snapshot-to-snapshot jitter within a single training run. For most audited metrics, training-seed and snapshot variance dominate: across three training seeds and ten post-warmup checkpoints, only seven of 34 metrics carry a between-variant share large enough to confidently distinguish from chance (the four core reconstruction metrics, sae-probes at $k\!\in\!\{1, 5\}$, and RAVEL disentangle; details in Appendix~\ref{app:multi_seed}). At SCR top-$\{5, 10\}$, TPP top-$\{10, 20, 100\}$, sparse probing top-$1$, sae-probes $k\!=\!16$, SCR top-$50$, and RAVEL cause, each of the three training seeds picks a different best variant among the four. A practitioner who runs these benchmarks once on a panel of similar SAEs reads a winner that is essentially seed-driven.

\paragraph{Cross-metric ranking agreement} On both panels, we compute pairwise Spearman $\rho$ of metric rankings across all SNR-informative metrics (SNR $\geq 1$, see Appendix~\ref{app:snr_full}). When SAEs are very different (cross-architecture panel) metrics broadly agree on the ranking ($\rho = +0.44$); when SAEs are similar (Matryoshka panel) metrics rank essentially independently ($\rho = +0.08$), with stark same-family disagreements such as default sparse probing $k$=1 vs $k$=10 negatively correlated at $\rho = -0.4$. Full heatmaps for both panels are in Appendix~\ref{app:contradiction}, Figure~\ref{fig:contradiction}.

\section{Validity: ground-truth correlation on synthetic SAEs}
\label{sec:gt}

A metric can have low reseed noise and good discriminability while still measuring the wrong thing: an arbitrary, repeatable function of SAE latents would pass both tests. To check that each metric's score actually reflects SAE quality, we need SAEs whose ground-truth quality is computable, which is only possible on synthetic models. We can then train SAEs on the synthetic model and check whether each proxy metric's score correlates with ground-truth quality. Sparse probing, TPP, and SCR can be validated this way. Autointerp requires real natural-language text, and RAVEL requires a real LLM to intervene on, so neither can be tested under this lens.

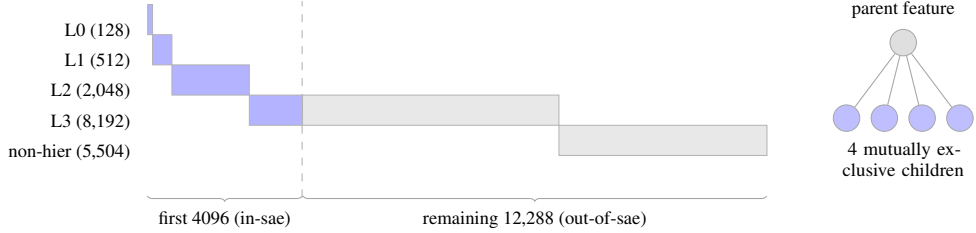
\begin{figure}[t]
\centering
\begin{tikzpicture}[
  font=\footnotesize,
  bar/.style={rectangle, draw=gray!60, line width=0.3pt, minimum height=4mm, inner sep=0, anchor=south west},
  inscope/.style={bar, fill=blue!30},
  outscope/.style={bar, fill=gray!18},
  brace/.style={decorate, decoration={brace, mirror, amplitude=2pt, raise=0pt}, line width=0.4pt, gray!70},
  task/.style={draw, rounded corners=2pt, inner xsep=5pt, inner ysep=4pt, align=left, font=\scriptsize, text width=3.85cm, anchor=north west, minimum height=2.4cm},
  arr/.style={-{Stealth[length=1.5mm]}, line width=0.4pt},
  dashedline/.style={dashed, gray!60, line width=0.4pt},
  llabel/.style={font=\scriptsize, anchor=east},
  treenode/.style={circle, draw=gray!70, fill=gray!25, minimum size=3.5mm, inner sep=0, line width=0.3pt},
  treechild/.style={circle, draw=gray!70, fill=blue!25, minimum size=3.5mm, inner sep=0, line width=0.3pt},
]

% --- Hierarchy staircase: each level offset to start where previous ended ---
% Scale: 1 feature = 0.0005 cm (so 16384 features span 8.192 cm)
% In-scope cutoff at index 4096 -> x = 2.048 cm
% Cumulative starts: L0=0, L1=128, L2=640, L3=2688, non-hier=10880

% L0: 128 root features, all in-sae (very narrow bar)
\node[llabel] at (-0.1cm, 4.45cm) {L0 (128)};
\node[inscope, minimum width=0.064cm] at (0, 4.4cm) {};

% L1: 512 features, all in-sae
\node[llabel] at (-0.1cm, 4.05cm) {L1 (512)};
\node[inscope, minimum width=0.256cm] at (0.064cm, 4.0cm) {};

% L2: 2048 features, all in-sae
\node[llabel] at (-0.1cm, 3.65cm) {L2 (2{,}048)};
\node[inscope, minimum width=1.024cm] at (0.32cm, 3.6cm) {};

% L3: 8192 features. First 1408 in-sae, remaining 6784 out-of-sae.
\node[llabel] at (-0.1cm, 3.25cm) {L3 (8{,}192)};
\node[inscope, minimum width=0.704cm] at (1.344cm, 3.2cm) {};
\node[outscope, minimum width=3.392cm] at (2.048cm, 3.2cm) {};

% non-hier: 5504 features, all out-of-sae
\node[llabel] at (-0.1cm, 2.85cm) {non-hier (5{,}504)};
\node[outscope, minimum width=2.752cm] at (5.44cm, 2.8cm) {};

% Vertical dashed line at the in-sae cutoff (index 4096, x = 2.048cm)
\draw[dashedline] (2.048cm, 2.25cm) -- (2.048cm, 4.85cm);

% Braces below the staircase
\draw[brace] (0, 2.27cm) -- (2.048cm, 2.27cm);
\node[font=\scriptsize, anchor=north] at (1.024cm, 2.19cm) {first 4096 (in-sae)};

\draw[brace] (2.06cm, 2.27cm) -- (8.192cm, 2.27cm);
\node[font=\scriptsize, anchor=north] at (5.126cm, 2.19cm) {remaining 12{,}288 (out-of-sae)};

% --- Right side: hierarchy tree (parent + 4 mutually exclusive children) ---
\node[treenode] (root) at (10cm, 4.3cm) {};
\node[treechild] (c1) at (9.25cm, 3.3cm) {};
\node[treechild] (c2) at (9.75cm, 3.3cm) {};
\node[treechild] (c3) at (10.25cm, 3.3cm) {};
\node[treechild] (c4) at (10.75cm, 3.3cm) {};
\draw[thin, gray!70] (root) -- (c1);
\draw[thin, gray!70] (root) -- (c2);
\draw[thin, gray!70] (root) -- (c3);
\draw[thin, gray!70] (root) -- (c4);
\node[font=\scriptsize, anchor=south] at (10cm, 4.5cm) {parent feature};
\node[font=\scriptsize, anchor=north, text width=2.6cm, align=center] at (10cm, 3.1cm) {4 mutually exclusive children};

\end{tikzpicture}
\caption{SynthSAEBench-16k contains a 16{,}384-feature ground-truth dictionary broken into hierarchy levels L0--L3, and non-hierarchical tail. We train SAEs of width 4096, so the first 4096 features of SynthSAEBench-16k are considered in-sae for the SAE (blue). The remaining 12{,}288 are out-of-sae (gray); the dashed line marks the cutoff, partway through L3.}
\label{fig:synth_tasks}
\end{figure}

\subsection{SynthSAEBench}

We use the SynthSAEBench~\citep{chanin2026synthsaebench} as a base for our evaluation. SynthSAEBench creates synthetic activations for SAE training with hierarchy, correlation, and superposition, simulating realistic phenomena in an LLM, while providing ground-truth features. The SynthSAEBench-16k model defines $16{,}384$ ground-truth feature directions $d_i \in \mathbb{R}^D$ as random unit vectors. Of these, $10{,}880$ are arranged in 128 disjoint trees of branching factor 4 and depth 3; the remaining $5{,}504$ are non-hierarchical. Within a tree, children are mutually exclusive and a child can fire only if its parent does. Hierarchical roots and non-hierarchical features fire independently with per-feature firing rates. 

To draw an activation, a binary firing pattern $z \in \{0, 1\}^{16{,}384}$ is sampled under these constraints, per-feature magnitudes are drawn from a rectified Gaussian $c_i = z_i \cdot \mathrm{ReLU}(\mu_i + \sigma_i \varepsilon_i)$ with $\varepsilon_i \sim \mathcal{N}(0, 1)$, and the SAE input is $x = D^\top c + b$, where $D$ stacks the unit-norm feature directions and $b$ is a bias. SAEs train on this activation pool and never see $D$. Feature firing probabilities follow a zipfian distribution, where earlier feature indices fire more frequently than later features.

SynthSAEBench provides two key metrics to score how well a trained SAE recovers the ground-truth features. \emph{Mean Correlation Coefficient (GT-MCC)} matches the SAE's decoder columns $w_j$ to the ground-truth directions on absolute cosine similarity (Hungarian assignment) and reports the mean cosine similarity of matched pairs; this measures how closely the SAE dictionary matches the ground-truth feature directions. \emph{GT-F1} matches each SAE latent to its best-correlated ground-truth feature and reads binary-classification F1 of latent firing against ground-truth feature firing across activations. This metric measures how well the SAE encoder fires the correct features at the correct times.

\subsection{Evaluation SAEs}

We train a series of SAEs of different architectures with width $d_{\text{sae}} = 4{,}096$ on SynthSAEBench, as recommended by SynthSAEBench~\citep{chanin2026synthsaebench}. We assume that the SAEs thus learn roughly the first 4{,}096 features of SynthSAEBench-16k. We call the first 4{,}096 SynthSAEBench features \emph{in-sae}, and the remaining 12{,}288 features \emph{out-of-sae}. This is demonstrated in Figure~\ref{fig:synth_tasks}.

\paragraph{Synthetic SAEs} We train 35 SAEs at $d_{\text{sae}} = 4096$. These SAEs use 5 architectures (standard, JumpReLU, BatchTopK, Matryoshka, Matching Pursuit) each at 7 L0 levels ($\{15, 20, 25, 30, 35, 40, 45\}$). Each SAE is trained on 200M samples following the SynthSAEBench recommended hyperparameters~\citep{chanin2026synthsaebench}.

\paragraph{Control SAEs} We include 3 degraded variants of the canonical JumpReLU L0=25 SAE: \texttt{random\_init} and \texttt{random\_l0\_matched} randomize weights, the latter with threshold tuned to L0$\,\approx 25$; \texttt{permuted\_decoder} shuffles $W_{\text{dec}}$ rows. We also include a \emph{perfect oracle} whose decoder is the first 4096 ground-truth feature directions and encoder returns exact ground-truth activations (GT-MCC = GT-F1 = 1.0 by construction).

\subsection{Synthetic task datasets}

\begin{table}[t]
\caption{Example tasks from the synthetic-task suite used in \S\ref{sec:gt}. Subscripts are ground-truth feature indices; in-sae means index $< 4{,}096$, out-of-sae means index $\geq 4{,}096$.}
\label{tab:synth_tasks}
\centering
\small
\begin{tabular}{ll}
\toprule
Task category & Example \\
\midrule
\multicolumn{2}{l}{\textit{Sparse probing} (binary classification on activations; train top-$k$ probe on SAE latents)} \\
\quad single in-sae        & is $f_{1{,}347}$ firing? \\
\quad single out-of-sae (negative control)   & is $f_{12{,}492}$ firing? \\
\quad boolean in-sae       & is $f_{17} \land f_{1{,}098}$ firing?  \quad or  \quad is $(f_{12} \land f_{123}) \lor f_{1{,}037}$ firing? \\
\quad boolean out-of-sae (negative control)   & is $f_{12{,}340} \lor f_{15{,}873}$ firing? \\
\quad boolean mixed        & is $f_{2{,}143} \land f_{15{,}873}$ firing?  \quad (one in-sae, one out-of-sae) \\
\midrule
\multicolumn{2}{l}{\textit{TPP} (4-sibling group at depth 3; ablate top-$N$ latents most damaging to focal probe)} \\
\quad in-sae group         & focal $f_{2{,}688}$, siblings $\{f_{2{,}689}, f_{2{,}690}, f_{2{,}691}\}$ \\
\quad out-of-sae group (negative control)    & focal $f_{8{,}000}$, siblings $\{f_{8{,}001}, f_{8{,}002}, f_{8{,}003}\}$ \\
\midrule
\multicolumn{2}{l}{\textit{SCR} ((T, S) pair; biased probe on $T \land S$, ablate top-$N$ latents associated with $S$)} \\
\quad in-sae pair          & target $T = f_{317}$, spurious $S = f_{1{,}422}$ (different L0 subtrees) \\
\bottomrule
\end{tabular}

\end{table}
 
We construct hierarchy-aligned synthetic analogues of each SAEBench benchmark, partitioned by whether the probing targets consist of feature indices that are in-sae, out-of-sae or a mix. We assume that human-interpretable concepts do not always map directly onto a single LLM feature (although many likely do). For instance, the concept ``is letter L'' might actually be represented as ``is capital L OR is lowercase L''. We simulate this with a boolean mix of ground-truth features.

All tasks for a given evaluation share a 60{,}000-activation pool drawn fresh from the synthetic model per task-generation seed (so the pool, the per-task labels, and the per-task feature selections all re-randomize on reseed); each task then subsets this pool to its own balanced positive/negative classes, with class-balanced positive/negative splits and an 80/20 train/test partition where applicable.

\paragraph{Sparse probing} We construct 92 binary classification tasks ("is dictionary feature $f$ firing in this activation?"), split by SAE coverage (in-sae / out-of-sae / mixed) and structure (single feature vs 2--3-feature boolean combination); for each task we draw a class-balanced split of up to 1{,}024 training activations (target 512 positives + 512 negatives) and up to 3{,}000 test activations (target 1{,}500 + 1{,}500), then train a $k$-sparse logistic regression on SAE latents and read top-$k$ accuracy.

\paragraph{TPP} We construct 60 four-sibling groups at depth 3 (analogue of SAEBench's 4-profession bias-in-bios setup), 30 in-sae and 30 out-of-sae; for each group we draw 2{,}000 samples per sibling class (up to 8{,}000 activations per group, balanced across the four siblings), train one probe per sibling under an 80/20 train/test split, identify the SAE latents whose ablation most damages the focal-sibling probe, and measure the gap between damage to the focal probe and damage to its three sibling probes.

\paragraph{SCR} We construct between 9 and 12 (T, S) pairs per seed (Appendix~\ref{app:scr_construction}) with features T and S from different hierarchy subtrees so they can co-occur. For each pair we draw 1{,}500 samples per class across the six SCR classes ($T$, $\neg T$, $S$, $\neg S$, $T \wedge S$, $\neg T \wedge \neg S$), balanced down to the smallest eligible class and split 80/20 train/test; we train a biased probe only on samples where T and S co-fire, then measure how much that bias is removed when the SAE latents most associated with S are ablated. All SCR pairs are in-sae because SCR requires the (T=1 $\wedge$ S=1) cell of the contingency table to be populated, and out-of-sae features fire too rarely for an out-of-sae control for SCR.

Table~\ref{tab:synth_tasks} gives one concrete example per task category. We include out-of-sae task variants as a negative control: if a proxy metric scores highly on a feature that the SAE is \textit{not explicitly tracking}, then something has gone awry.

\begin{figure}[t]
\centering
\includegraphics[width=\linewidth]{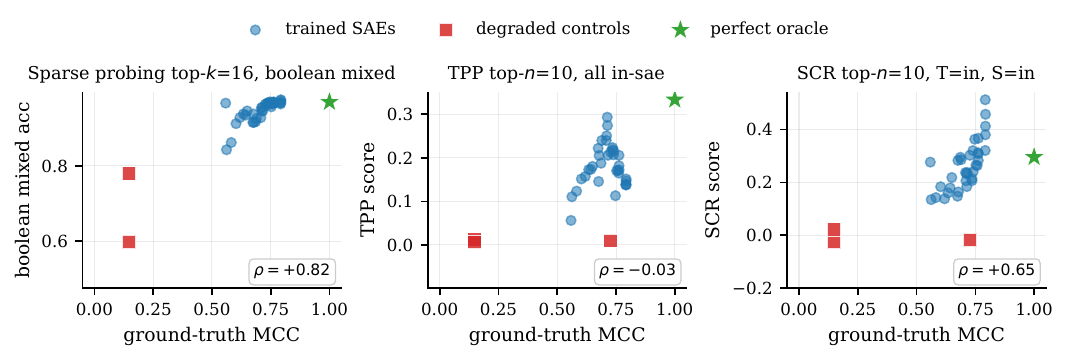}
\caption{Benchmark score vs GT-MCC at canonical hyperparameters, each point is one SAE. Sparse probing (left) uses the boolean-mixed task category. TPP (middle) and SCR (right) use canonical top-$N$=10. Sparse probing is well-calibrated with GT-MCC. TPP shows no correlation with GT-MCC. Problematically, SCR gives a lower score to the perfect oracle than to several of the trained SAEs.}
\label{fig:validity_scatter}
\end{figure}

\begin{table}[t]
\caption{Spearman $\rho$ between each benchmark's score and ground-truth quality metric. Metrics should correlate with either GT-MCC or GT-F1 for in-sae and mixed tasks, and should not correlate with out-of-sae tasks. Sparse probing is well-calibrated to GT-MCC across $k$. SCR becomes negatively correlated with GT-MCC at large top-$N$.}
\label{tab:realsae_correlations}
\centering
\small
\setlength{\tabcolsep}{4pt}
\begin{tabular}{l ccc @{\hspace{1.5em}} c @{\hspace{1.5em}} ccc}
\toprule
                                    & \multicolumn{3}{c}{$\rho$ vs GT-MCC} & & \multicolumn{3}{c}{$\rho$ vs GT-F1} \\
\cmidrule(lr){2-4} \cmidrule(lr){6-8}
\textit{Sparse probing, top-$k$ =}  & \textit{1}  & \textit{4}   & \textit{16}  & & \textit{1}  & \textit{4}   & \textit{16}  \\
\quad single in-sae                 & $+0.70$ & $+0.76$ & $+0.49$ & & $+0.31$ & $+0.46$ & $-0.22$ \\
\quad boolean in-sae                & $+0.48$ & $+0.68$ & $+0.87$ & & $+0.47$ & $+0.37$ & $+0.51$ \\
\quad boolean mixed                 & $+0.55$ & $+0.65$ & $+0.82$ & & $+0.52$ & $+0.45$ & $+0.46$ \\
\arrayrulecolor{black!35}\midrule[0.3pt]\arrayrulecolor{black}
\quad single out-of-sae (negative control)            & $+0.01$ & $+0.23$ & $+0.19$ & & $-0.35$ & $-0.58$ & $-0.37$ \\
\quad boolean out-of-sae (negative control)           & $-0.29$ & $-0.41$ & $-0.22$ & & $-0.25$ & $-0.53$ & $-0.53$ \\
\midrule
\textit{TPP, top-$N$ =}             & \textit{10} & \textit{100} & \textit{500} & & \textit{10} & \textit{100} & \textit{500} \\
\quad all in-sae                    & $-0.03$ & $+0.46$ & $+0.60$ & & $-0.69$ & $-0.57$ & $-0.49$ \\
\arrayrulecolor{black!35}\midrule[0.3pt]\arrayrulecolor{black}
\quad all out-of-sae (negative control)               & $+0.42$ & $+0.75$ & $+0.71$ & & $-0.03$ & $+0.36$ & $+0.25$ \\
\midrule
\textit{SCR, top-$N$ =}             & \textit{10} & \textit{100} & \textit{500} & & \textit{10} & \textit{100} & \textit{500} \\
\quad $T$=in, $S$=in                & $+0.65$ & $-0.15$ & $-0.64$ & & $+0.41$ & $-0.34$ & $-0.42$ \\
\bottomrule
\end{tabular}
\end{table}

\subsection{Results}
\label{sec:results}

For each proxy metric task, we calculate the Spearman correlation between the proxy metric and the ground-truth metrics GT-MCC and GT-F1. We consider it a success if the proxy metric tracks either metric, as the ground-truth metrics themselves track different quality aspects of the underlying SAE, and do not always peak at the same L0. We report correlation results for all tasks in Table~\ref{tab:realsae_correlations}. We show a subset of scatter plots for each task in Figure~\ref{fig:validity_scatter}. We additionally run ablation experiments on modified versions of the SynthSAEBench-16k base model in Appendix~\ref{app:variations}, which probes whether the calibration we observe is a property of the v1 base model or holds across feature distributions.

\paragraph{Sparse probing is reasonably well calibrated to GT-MCC} Sparse probing interestingly tracks GT-MCC better than GT-F1, despite being an encoder-only metric. Sparse probing does not penalize latents for firing more than they should since the $k$-sparse probe can learn its own threshold per firing latent. GT-MCC stays high at high L0, while GT-F1 decays past a clear optimal L0 on most architectures (Appendix~\ref{app:gt_vs_l0}, Figure~\ref{fig:gt_vs_l0}). This is not necessarily a problem, as GT-MCC is a valid metric.

One potential problem with sparse-probing is that it saturates well before perfect GT-MCC. At canonical top-$k=16$, the in-sae single-feature accuracy is above $0.99$ for almost every trained SAE in this study, while their GT-MCC values span the much wider range $0.56$--$0.79$. Without spread on the metric side, sparse probing struggles to differentiate between high-quality SAEs.

% \textbf{Calibration is firing-rate-dependent.} On higher-firing-rate variations (\texttt{rel-p-1.5}, \texttt{std-2.5}), single out-of-sae $\rho$ rises to $+0.57$--$+0.63$, undermining the agnosticism above; the lower-firing-rate \texttt{rel-p-0.5} preserves it ($\rho = +0.06$). Mechanism: when out-of-sae features fire often, hidden-space proxies for them become accessible to a probe via correlated in-sae latents. Sparse probing's calibration on absent features is thus a property of v1's firing-rate regime, not a guarantee.

\paragraph{TPP is poorly correlated with ground-truth} TPP correlates poorly with GT-MCC at low top-$N$, with essentially 0 correlation at the canonical top-$N=10$. Strangely, TPP appears to be better-correlated with out-of-sae features on SynthSAEBench-16k, implying it is tracking some sort of feature mixing between in-sae and out-of-sae features rather than directly measuring SAE quality, but this also varies a lot in our variant model ablations (Appendix~\ref{app:variations}). Either way, the canonical TPP score is not a reliable proxy for SAE quality on this synthetic panel.

\paragraph{SCR is unreliable across multiple axes} SCR has three serious problems in this synthetic test. First, even at the canonical top-$N=10$, SCR ranks the perfect oracle lower than 11 of the 35 trained SAEs tested. Second, when top-$N$ is 50 or higher, the metric becomes \emph{negatively correlated} with all ground-truth metrics ($\rho_{\text{MCC}} = -0.31$ at top-$N=50$, $-0.64$ at top-$N=500$ on the v1 panel; Table~\ref{tab:realsae_correlations}), meaning better SAEs score lower on SCR than worse SAEs. We document further SCR failures on variant synthetic models in Appendix~\ref{app:variations}, Table~\ref{tab:variations}.

\section{Related work}

\paragraph{SAE benchmarking failures.} %\citet{heap2025sparse} show that several common SAE metrics fail to clearly separate SAEs trained on randomly initialized transformers from SAEs trained on real ones.
\citet{paulo2025sparse} find that only $\sim$30\% of features are shared across seed-replicated training runs of the same SAE, raising identifiability concerns. \citet{korznikov2026sanity} report that random-feature baselines match trained SAEs on autointerp, sparse probing, and causal editing; \citet{song2025position} propose a feature-consistency metric (PW-MCC) as a complementary reliability axis. Our audit is complementary to these: rather than introducing a single new diagnostic, we evaluate the existing SAEBench suite along three orthogonal lenses (reseed noise, ground-truth correlation, and discriminability across training).

\paragraph{Cases where supervised baselines beat SAEs.} \citet{kantamneni2025sparse} report that SAE-based sparse probes underperform plain logistic regression across data scarcity, label noise, and covariate shift; \citet{wu2025axbench} find SAE steering inferior to prompting and difference-in-means representations. DeepMind's recent negative-results report~\citep{smith2025negative} echoes both findings on a safety-relevant probing task. These results suggest the existing proxy metrics overstate SAE quality on downstream tasks, motivating an audit of the proxies themselves.

\paragraph{Ground-truth and synthetic SAE evaluation.} The toy-model lineage initiated by \citet{elhage2022toy} establishes sparse linear combinations of orthogonal feature directions as a controlled testbed for dictionary learning. \citet{karvonen2024measuring} run an analogous evaluation on Othello and chess board states, and \citet{makelov2024towards} construct supervised dictionaries on the IOI task as a comparison ground truth. \citet{oneill2025compute} introduce the optimal-bipartite-matching MCC metric that SynthSAEBench~\citep{chanin2026synthsaebench} adopts. We use SynthSAEBench-16k as our validity panel.

\paragraph{Benchmark auditing in ML more broadly.} \citet{liao2021learning} catalog 107 surveys of evaluation failures, distinguishing internal validity (noise, leakage, baselines) from external validity (transfer of progress to other tasks); \citet{bowman2021will} propose explicit desiderata for NLU benchmarks, the format we follow. \citet{dehghani2021benchmark} show that re-aggregating different subsets of SuperGLUE substantially reorders rankings of competing models. \citet{henderson2018deep} and \citet{reimers2017reporting} study on seed variance in deep RL and NLP. \citet{card2020little} quantifies when test-set sizes are too small for benchmark comparisons to have statistical power.

\section{Discussion}
\label{sec:forward}

\paragraph{Practitioners should avoid SCR and TPP} TPP and SCR fail multiple basic sanity checks at their canonical settings on trained SAEs. TPP %correlates with quality on features the SAE does not even represent (\S\ref{sec:results}, Table~\ref{tab:realsae_correlations}) and
declines through training at top-$N \geq 50$ across every SAE we tested, and is poorly correlated with ground truth in our synthetic tests. SCR is negatively correlated with ground-truth at top-$N \geq 50$, declines through training at canonical top-$N=10$, and swings from $\rho=+0.90$ to $\rho=-0.10$ against GT-MCC across reasonable variations of the same base-model architecture (\S\ref{sec:results}, Appendix~\ref{app:variations}). We recommend avoiding SCR and TPP benchmarks.

% \paragraph{SCR and TPP are quality metrics, in the sense that matters} A defender of SCR or TPP might argue that they measure specific properties (spurious-correlation removal, targeted-probe perturbation effect) rather than overall SAE quality, and so should not be expected to track ground-truth feature recovery. We disagree with this defense as the metrics are currently configured. SCR and TPP are included in SAEBench's quality-metric suite and reported alongside metrics like sparse probing in standard SAEBench output; published comparisons of SAE architectures use them as quality scores. If they measure something other than SAE quality, they should be removed from the quality suite and re-framed as diagnostic tools. As currently configured and reported, they are quality metrics, and they fail the desiderata of quality metrics.

% \subsection{Toward better SAE benchmarks}
% \label{sec:better_benchmarks}

\paragraph{SAE metrics struggle with noise and discriminability} Many metrics have inter-checkpoint jitter comparable to their inter-variant signal, making them hard to use as a clear target for the field to hill-climb on. The most promising directions for future work are increasing the number and variety of datasets used by each metric, improving the reliability of internal probing steps, and replacing random latent sampling with reproducible selections that transfer across SAE training runs.

\paragraph{Limitations of our audit} Even for metrics that pass our checks, we cannot say with certainty that a proxy metric is ``correct'' on LLMs. It is possible for biases or dataset limitations to remain that we cannot measure without ground-truth on real LLMs. Limitations are discussed further in Appendix~\ref{app:limitations}.

\paragraph{The field needs better SAE benchmarks} Without clear and reliable benchmarks, the field will struggle to improve SAEs. Progress on SAE architectures requires stacking small improvements that the current suite cannot reliably detect. We believe SAE benchmarking is solvable, and we hope this audit catalyzes further work to extend and improve the existing suite.

\begin{ack}
David Chanin was supported thanks to EPSRC EP/S021566/1 and the Machine Learning Alignment and Theory Scholars (MATS) program. We are grateful to Fred Bruford for feedback during this project.
\end{ack}

\bibliographystyle{plainnat}
\bibliography{references}

\appendix

\section{Limitations}
\label{app:limitations}

The synthetic-SAE correlation analysis (\S\ref{sec:gt}) on SynthSAEBench-16k uses three task-generation seeds; the SAE-variation models are evaluated at a single seed each, so the cross-base-model fragility of SCR canonical we report combines genuine cross-model variation with single-seed task-sampling noise. The cross-architecture training-trajectory analysis (\S\ref{sec:snr}) uses a single training seed per variant, so per-step jitter on that panel is a lower bound on the metric's pure noise (it includes some actual SAE-quality drift across snapshots); the sampled-Matryoshka panel uses three training seeds per variant and is decomposed in Appendix~\ref{app:multi_seed}. Three seeds is the minimum that supports a variance decomposition, so individual share values there are imprecise even though the qualitative ordering is robust. Our discriminability (\S\ref{sec:snr}) and validity (\S\ref{sec:gt}) analyses use a single underlying language model (Gemma-2-2b); discriminability and validity may behave differently on other model families, though the four-SAE cross-model reseed-noise comparison in Appendix~\ref{app:per_sae_noise} suggests the noise patterns themselves are largely model-agnostic. The four-SAE sampled-Matryoshka panel covers 300M tokens of training, so trajectories may not have converged. SynthSAEBench-16k is one synthetic model architecture, so generalizing the calibration findings to other ground-truth setups is open work.

We do not evaluate the following SAEBench metrics: \emph{absorption}~\citep{chanin2024absorption} and \emph{meta\_structure} diagnose specific SAE failure modes (feature absorption and latent-space meta-structure) rather than scoring quality, and \emph{unlearning} targets instruct-tuned-model SAEs while every other SAEBench evaluation and the SAE training literature target base models. Several audited evaluations originated as standalone benchmarks~\citep{gurnee2023finding,bills2023language,kantamneni2025sparse,marks2025sparse,huang2024ravel}; we audit their SAEBench-as-configured versions only. We also consider AxBench~\citep{wu2025axbench} as out of scope, as it is a general steering benchmark rather than being SAE-specific.

\section{SAEBench metric details}
\label{app:metric_details}

This appendix gives the full procedure for each of the six SAEBench quality metrics audited in the paper. The condensed descriptions in \S\ref{sec:background} are sufficient for following the audit findings; this appendix is the reference for readers who want the implementation-level setup, hyperparameters, and dataset choices.

\paragraph{Sparse probing}~\citep{gurnee2023finding} measures whether SAE latents disentangle natural-language concepts into a small number of latents. For each binary probing task (e.g.\ ``is this text in French'' or ``is this profession a doctor''), SAEBench encodes the text into SAE latents, identifies the top-$k$ latents most informative about the binary label using per-latent statistics, and trains a logistic-regression probe restricted to those $k$ latents. The headline metric is test accuracy at $k \in \{1, 2, 5\}$; higher accuracy at small $k$ indicates the relevant concept is concentrated in a small number of latents. The default benchmark uses 5 probing datasets covering language identification, profession (bias-in-bios), and topic classification.

\paragraph{Sae-probes sparse probing}~\citep{kantamneni2025sparse} is a variant of sparse probing that runs the same conceptual evaluation on a much larger and more diverse task suite. Sae-probes uses 113 binary classification tasks (vs sparse probing's 5), trains an L1-regularized logistic-regression probe with cross-validation over the L1 strength, and reports test AUC, F1, and accuracy at multiple values of $k$. The cross-validated regularization makes sae-probes results less sensitive to the specific probe-training hyperparameters, and the larger task suite reduces the dataset-sampling noise of the headline number.

\paragraph{Spurious Correlation Removal (SCR)}~\citep{marks2025sparse} tests whether SAE latents can be used to remove a spurious correlation from a probe without retraining. The setup picks a target attribute $T$ (e.g.\ profession) and a spurious attribute $S$ (e.g.\ gender) and trains a logistic-regression probe to predict $T$ on a confounded subset of the data containing only $T \wedge S$ and $\neg T \wedge \neg S$ examples, so the probe can solve the training task by reading either attribute. SCR then identifies the top-$N$ SAE latents most associated with $S$ (by largest mean-activation difference between $S\!=\!1$ and $S\!=\!0$ examples), ablates them, and reports how much of the accuracy gap between this biased probe and an oracle probe trained on un-confounded data is closed on a held-out balanced test set where $T$ and $S$ are independent. A higher SCR score means the SAE successfully isolated the spurious attribute into a small set of removable latents. The benchmark uses 2 datasets, and the canonical top-$N$ is 10; SAEBench's default sweep reports top-$N \in \{2, 5, 10, 20, 50, 100, 500\}$. The internal probe is trained with $\ell_1$ penalty $10^{-3}$ at batch size 16; both are load-bearing for the metric value (Appendix~\ref{app:scr_batch}).

\paragraph{Targeted Probe Perturbation (TPP)}~\citep{marks2025sparse} measures whether the SAE causally separates concepts within a single category. The default setup uses bias-in-bios restricted to a 4-class group of professions. For each focal class $c$ in the group, SAEBench trains four binary probes (one per class) on the SAE latents, then identifies the top-$N$ latents whose ablation most damages the focal probe's accuracy, zeroes those $N$ latents, and measures the accuracy drop on the focal probe versus the average accuracy drop on the three non-focal probes. The TPP score is the gap (focal drop minus non-focal drop), averaged over choice of focal class. A high TPP means each class's representation can be isolated to a small set of latents that does not affect the other classes' probes. The benchmark uses 2 datasets and reports across $N \in \{2, 5, 10, 20, 50, 100, 500\}$. As with SCR, the internal probe-training $\ell_1$ penalty and batch size matter for the metric value.

\paragraph{Autointerp}~\citep{bills2023language,paulo2024automatically} measures how well SAE latents can be described in natural language and how well those descriptions predict latent firing. SAEBench's workflow first selects $N$ latents from the SAE (uniformly at random by default), collects the top-activating natural-language examples for each, asks an LLM (the explainer) to generate a natural-language description of what each latent fires on, and then asks a second LLM (the scorer) to predict, given the description, which held-out examples should activate the latent. The autointerp score is the agreement between scorer predictions and ground-truth firing patterns. The default uses $N=1000$ randomly-selected latents and a total token budget of 2M. We additionally evaluate a fixed-latent variant where the same 1000 latents are used across reseeds (selected via seed 42), so reseed noise is isolated to the LLM-judge variance and excludes latent-sampling noise.

\paragraph{RAVEL}~\citep{huang2024ravel} measures whether SAE latents can be used to causally edit one attribute of an entity without affecting other attributes. The setup uses entity-attribute datasets (e.g.\ cities with attributes country, language, continent), and for each (target attribute, distractor attribute) pair the RAVEL pipeline finds SAE latents whose ablation changes the model's prediction of the target attribute (the \emph{cause} score) without changing its prediction of the distractor attribute (the \emph{disentangle} score); the \emph{isolation} score combines the two. SAEBench reports all three sub-scores; in the body we focus on \emph{disentangle}, which has the lowest reseed noise of the three. RAVEL's default datasets are Gemma-2-2b specific (entity sets and prompt templates are hard-coded for that base model), which is why we restrict the RAVEL evaluation to Gemma Scope SAEs in Appendix~\ref{app:per_sae_noise}.

\section{Per-SAE reseed noise tables}
\label{app:per_sae_noise}

Tables~\ref{tab:noise_full_65k}, \ref{tab:noise_full_16k}, \ref{tab:noise_full_gpt2}, and \ref{tab:noise_full_llama} give the full per-benchmark breakdown of reseed CV on a four-SAE panel spanning three model families: the canonical 65k and 16k Gemma Scope SAEs (residual stream layer 12 of Gemma-2-2b), the canonical gpt2-small 32k SAE (gpt2-small layer 6 \texttt{hook\_resid\_post}), and the canonical Llama Scope 8x SAE (Llama-3.1-8B layer 16 residual, width 32k). All four use five reseeds at seeds $\{42, 123, 456, 789, 2024\}$ at canonical hyperparameters. Same protocol as \S\ref{sec:noise}. Table~\ref{tab:noise} in \S\ref{sec:noise} is the curated subset of Table~\ref{tab:noise_full_65k}. The qualitative pattern is the same across all four SAEs: sae-probes, sparse probing top-$k$, RAVEL disentangle, and autointerp are sub-1\% CV; TPP and SCR at large top-$N$ are mid-single-digit CV (with the notable exception of SCR top-500 on gpt2-small at 27\%); TPP at small top-$N$ has very high relative noise because the score itself is close to zero. RAVEL is restricted to the two Gemma SAEs because the SAEBench RAVEL wrapper hard-codes datasets for Gemma-2-2b only.

\begin{table}[h]
\caption{Full reseed noise breakdown on the canonical 65k Gemma Scope SAE, five reseeds.}
\label{tab:noise_full_65k}
\centering
\small
\begin{tabular}{lrrr}
\toprule
Benchmark & Mean & Std & CV \\
\midrule
sae-probes ($k=1$, accuracy)             & 0.763 & 0.001 & 0.2\% \\
sae-probes ($k=2$, accuracy)             & 0.780 & 0.001 & 0.1\% \\
sae-probes ($k=5$, accuracy)             & 0.800 & 0.002 & 0.2\% \\
sparse\_probing ($k=1$, accuracy)        & 0.722 & 0.009 & 1.2\% \\
sparse\_probing ($k=2$, accuracy)        & 0.773 & 0.005 & 0.6\% \\
sparse\_probing ($k=5$, accuracy)        & 0.852 & 0.003 & 0.3\% \\
RAVEL cause                               & 0.676 & 0.019 & 2.8\% \\
RAVEL disentangle                         & 0.706 & 0.001 & 0.2\% \\
RAVEL isolation                           & 0.736 & 0.018 & 2.5\% \\
autointerp random-latent ($n_{\text{latents}}=1000$) & 0.839 & 0.004 & 0.5\% \\
autointerp fixed-latent ($n_{\text{latents}}=1000$)  & 0.846 & 0.002 & 0.2\% \\
SCR top-2                                  & 0.048 & 0.003 & 6.4\% \\
SCR top-5                                  & 0.114 & 0.005 & 4.4\% \\
SCR top-10                                 & 0.174 & 0.008 & 4.4\% \\
SCR top-20                                 & 0.263 & 0.008 & 3.2\% \\
SCR top-50                                 & 0.349 & 0.016 & 4.7\% \\
SCR top-100                                & 0.364 & 0.016 & 4.5\% \\
SCR top-500                                & 0.351 & 0.022 & 6.2\% \\
TPP top-2                                  & 0.005 & 0.001 & 19\% \\
TPP top-5                                  & 0.010 & 0.002 & 16\% \\
TPP top-10                                 & 0.025 & 0.010 & 39\% \\
TPP top-20                                 & 0.045 & 0.016 & 36\% \\
TPP top-50                                 & 0.094 & 0.021 & 23\% \\
TPP top-100                                & 0.157 & 0.022 & 14\% \\
TPP top-500                                & 0.345 & 0.010 & 2.8\% \\
\bottomrule
\end{tabular}

\end{table}

\begin{table}[h]
\caption{Full reseed noise breakdown on the canonical 16k Gemma Scope SAE, five reseeds.}
\label{tab:noise_full_16k}
\centering
\small
\begin{tabular}{lrrr}
\toprule
Benchmark & Mean & Std & CV \\
\midrule
sae-probes ($k=1$, accuracy)             & 0.763 & 0.001 & 0.1\% \\
sae-probes ($k=2$, accuracy)             & 0.782 & 0.001 & 0.1\% \\
sae-probes ($k=5$, accuracy)             & 0.806 & 0.001 & 0.2\% \\
sparse\_probing ($k=1$, accuracy)        & 0.756 & 0.004 & 0.5\% \\
sparse\_probing ($k=2$, accuracy)        & 0.804 & 0.003 & 0.3\% \\
sparse\_probing ($k=5$, accuracy)        & 0.871 & 0.004 & 0.4\% \\
RAVEL cause                               & 0.655 & 0.008 & 1.2\% \\
RAVEL disentangle                         & 0.696 & 0.003 & 0.4\% \\
RAVEL isolation                           & 0.736 & 0.011 & 1.4\% \\
autointerp random-latent ($n_{\text{latents}}=1000$) & 0.820 & 0.003 & 0.4\% \\
autointerp fixed-latent ($n_{\text{latents}}=1000$)  & 0.829 & 0.003 & 0.4\% \\
SCR top-2                                  & 0.132 & 0.004 & 3.0\% \\
SCR top-5                                  & 0.215 & 0.008 & 3.6\% \\
SCR top-10                                 & 0.287 & 0.009 & 3.0\% \\
SCR top-20                                 & 0.363 & 0.011 & 3.1\% \\
SCR top-50                                 & 0.410 & 0.022 & 5.3\% \\
SCR top-100                                & 0.347 & 0.021 & 6.2\% \\
SCR top-500                                & 0.302 & 0.037 & 12\% \\
TPP top-2                                  & 0.009 & 0.001 & 14\% \\
TPP top-5                                  & 0.022 & 0.009 & 42\% \\
TPP top-10                                 & 0.037 & 0.015 & 39\% \\
TPP top-20                                 & 0.069 & 0.021 & 31\% \\
TPP top-50                                 & 0.138 & 0.025 & 18\% \\
TPP top-100                                & 0.220 & 0.024 & 11\% \\
TPP top-500                                & 0.400 & 0.006 & 1.6\% \\
\bottomrule
\end{tabular}

\end{table}

\begin{table}[h]
\caption{Full reseed noise breakdown on the canonical gpt2-small 32k SAE (gpt2-small layer 6 \texttt{hook\_resid\_post}), five reseeds. RAVEL is omitted because the SAEBench RAVEL wrapper has no gpt2 dataset.}
\label{tab:noise_full_gpt2}
\centering
\small
\begin{tabular}{lrrr}
\toprule
Benchmark & Mean & Std & CV \\
\midrule
sae-probes ($k=1$, accuracy)             & 0.690 & 0.001 & 0.2\% \\
sae-probes ($k=2$, accuracy)             & 0.704 & 0.001 & 0.1\% \\
sae-probes ($k=5$, accuracy)             & 0.724 & 0.001 & 0.2\% \\
sparse\_probing ($k=1$, accuracy)        & 0.700 & 0.006 & 0.9\% \\
sparse\_probing ($k=2$, accuracy)        & 0.776 & 0.003 & 0.4\% \\
sparse\_probing ($k=5$, accuracy)        & 0.822 & 0.003 & 0.4\% \\
autointerp random-latent ($n_{\text{latents}}=1000$) & 0.868 & 0.003 & 0.4\% \\
autointerp fixed-latent ($n_{\text{latents}}=1000$)  & 0.867 & 0.002 & 0.3\% \\
SCR top-2                                  & 0.064 & 0.004 & 6.8\% \\
SCR top-5                                  & 0.092 & 0.005 & 5.8\% \\
SCR top-10                                 & 0.128 & 0.005 & 4.2\% \\
SCR top-20                                 & 0.181 & 0.004 & 2.2\% \\
SCR top-50                                 & 0.231 & 0.008 & 3.3\% \\
SCR top-100                                & 0.250 & 0.008 & 3.3\% \\
SCR top-500                                & 0.222 & 0.060 & 27\% \\
TPP top-2                                  & 0.005 & 0.002 & 29\% \\
TPP top-5                                  & 0.012 & 0.003 & 25\% \\
TPP top-10                                 & 0.021 & 0.003 & 14\% \\
TPP top-20                                 & 0.039 & 0.001 & 3.5\% \\
TPP top-50                                 & 0.088 & 0.006 & 7.3\% \\
TPP top-100                                & 0.147 & 0.009 & 6.0\% \\
TPP top-500                                & 0.337 & 0.009 & 2.6\% \\
\bottomrule
\end{tabular}

\end{table}

\begin{table}[h]
\caption{Full reseed noise breakdown on the canonical Llama Scope 8x SAE (Llama-3.1-8B layer 16 residual, width 32k), five reseeds. RAVEL is omitted because the SAEBench RAVEL wrapper has no Llama dataset.}
\label{tab:noise_full_llama}
\centering
\small
\begin{tabular}{lrrr}
\toprule
Benchmark & Mean & Std & CV \\
\midrule
sae-probes ($k=1$, accuracy)             & 0.742 & 0.003 & 0.4\% \\
sae-probes ($k=2$, accuracy)             & 0.758 & 0.001 & 0.1\% \\
sae-probes ($k=5$, accuracy)             & 0.785 & 0.003 & 0.3\% \\
sparse\_probing ($k=1$, accuracy)        & 0.782 & 0.003 & 0.4\% \\
sparse\_probing ($k=2$, accuracy)        & 0.825 & 0.004 & 0.5\% \\
sparse\_probing ($k=5$, accuracy)        & 0.869 & 0.004 & 0.4\% \\
autointerp random-latent ($n_{\text{latents}}=1000$) & 0.860 & 0.004 & 0.4\% \\
autointerp fixed-latent ($n_{\text{latents}}=1000$)  & 0.858 & 0.003 & 0.3\% \\
SCR top-2                                  & 0.045 & 0.005 & 11\% \\
SCR top-5                                  & 0.079 & 0.007 & 9.0\% \\
SCR top-10                                 & 0.121 & 0.006 & 4.6\% \\
SCR top-20                                 & 0.163 & 0.005 & 3.2\% \\
SCR top-50                                 & 0.182 & 0.005 & 2.8\% \\
SCR top-100                                & 0.227 & 0.007 & 3.1\% \\
SCR top-500                                & 0.241 & 0.007 & 2.8\% \\
TPP top-2                                  & 0.008 & 0.002 & 20\% \\
TPP top-5                                  & 0.013 & 0.002 & 14\% \\
TPP top-10                                 & 0.017 & 0.003 & 18\% \\
TPP top-20                                 & 0.026 & 0.004 & 14\% \\
TPP top-50                                 & 0.045 & 0.003 & 5.7\% \\
TPP top-100                                & 0.062 & 0.003 & 4.3\% \\
TPP top-500                                & 0.127 & 0.004 & 2.9\% \\
\bottomrule
\end{tabular}

\end{table}

\section{Derivation of single-seed comparison thresholds}
\label{app:threshold_derivation}

The fifth column of Table~\ref{tab:noise} reports a per-metric threshold $|\Delta|^*$ for a 95\% two-tailed comparison of two single-seed scores. We derive it here.

\paragraph{The test.} If a practitioner evaluates two SAEs at one seed each and compares their scores $s_A$ and $s_B$, the difference $\Delta = s_A - s_B$ has variance $2\sigma^2$ under independent reseeds, where $\sigma$ is the metric's per-seed standard deviation. We do not know $\sigma$ exactly: the column-3 std $s$ in Table~\ref{tab:noise} is estimated from only $n=5$ canonical-SAE reseeds, so the right test is a two-sample $t$-test rather than a $z$-test. Under H$_0$ ``the two SAEs have equal true score'' the statistic $\Delta/(s\sqrt{2})$ follows a $t$-distribution with $n-1=4$ degrees of freedom; the 95\% two-tailed threshold is therefore
\[
|\Delta|^* \;=\; t_{0.025,\,4}\cdot s\sqrt{2} \;\approx\; 3.93\,s.
\]
The equivalent multiplier under known $\sigma$ would be $1.96\sqrt{2}\approx 2.77$; the inflation factor $t_{0.025,4}/z_{0.025}\approx 1.41$ reflects the chi-squared uncertainty in $s$ given only 5 reseeds (the 95\% CI on $\sigma$ from $n=5$ samples is roughly $0.6s$ to $2.9s$). Concretely: a practitioner who runs sae-probes on two SAEs and sees a $0.005$ gap should not conclude one is better, because the threshold is $0.008$; on SCR top-$500$ the same logic requires $|\Delta| > 0.09$ before a single-seed comparison is informative.

\paragraph{Multi-seed extension.} Averaging $S$ independent reseeds of each SAE shrinks the per-mean variance from $\sigma^2$ to $\sigma^2/S$, so $\mathrm{Var}(\Delta) = 2\sigma^2/S$ and the threshold scales as $|\Delta|^*_S = t_{0.025,\,2S-2}\cdot s\sqrt{2/S}$, where the larger pooled $df=2S-2$ also shrinks the multiplier toward the $z$ limit. For $S=3$ (e.g.\ the Matryoshka-panel design of \S\ref{sec:snr}) the threshold is roughly $1.6\,s$; with $S=5$ it is roughly $1.0\,s$.

\paragraph{Cross-SAE applicability.} Column $|\Delta|^*$ is derived from reseed noise on the canonical 65k Gemma Scope SAE and applies most directly to comparisons involving that SAE. Reseed std varies up to $\sim$2$\times$ across the four canonical SAEs in our cross-model panel (Tables~\ref{tab:noise_full_65k}--\ref{tab:noise_full_llama})~--- e.g.\ SCR top-$10$ std is $0.005$ on gpt2-small but $0.009$ on the 16k Gemma Scope SAE, $\sim$1.8$\times$ apart. We recommend treating Table~\ref{tab:noise}'s column 5 as a useful single-figure rule of thumb but, when a more careful threshold is needed for a specific SAE, recomputing $|\Delta|^*$ from that SAE's own per-metric std using the formula above. The full per-SAE std breakdowns are in Appendix~\ref{app:per_sae_noise}.

\section{Diagnostic-SAE \texorpdfstring{$\times$}{x} six-benchmark validation}
\label{app:diagnostic_grid}

We applied the two real-LLM diagnostic SAEs introduced in \S\ref{sec:gt} (\texttt{permuted\_decoder} and \texttt{random\_l0\_matched}) to all six SAEBench evaluations on the canonical Gemma-Scope JumpReLU SAE (Gemma-2-2b layer 12 residual, width 16k, mean L0 $124.7$); both were rebuilt for this SAE. Each diagnostic was run at 3 reseeds per cell with activation caching disabled, matching the reseed-noise protocol of \S\ref{sec:noise}. \texttt{random\_l0\_matched} achieved L0 $124.5$ via a uniform threshold tuned on a 20k-token Gemma-2-2b L12 calibration set.

\paragraph{Reduced evaluation configuration.} For compute reasons the diagnostic-SAE evaluation that produced Table~\ref{tab:diagnostic_grid} used a smaller subset of the SAEBench default configuration: sparse probing on 3 of the 8 default datasets (\texttt{LabHC/bias\_in\_bios\_class\_set1}, \texttt{canrager/amazon\_reviews\_mcauley\_1and5}, \texttt{fancyzhx/ag\_news}); SCR on the single \texttt{LabHC/bias\_in\_bios\_class\_set1} dataset (4 of the canonical 8 sub-tasks); and autointerp at $N=200$ randomly-sampled latents instead of the canonical $N=1000$. All three rows of Table~\ref{tab:diagnostic_grid} use this same reduced configuration, so the diagnostic comparisons (canonical vs each degraded variant) are internally valid. As a consequence of the reduced configuration, absolute scores in Table~\ref{tab:diagnostic_grid} differ from those in Appendix~\ref{app:per_sae_noise} (5 datasets, $N\!=\!1000$) and Appendix~\ref{app:hf_baselines} (canonical configuration); for example, canonical SCR top-10 is $0.208$ here versus $0.287$ in the canonical-config tables. The qualitative diagnostic conclusions reported below depend on relative comparisons within the table and are unaffected by the configuration choice.

\begin{table}[h]
\caption{Diagnostic SAE scores across the six SAEBench benchmarks under the reduced configuration described above, all on the canonical Gemma-Scope SAE. Headline metric per benchmark: \texttt{sae\_top\_5\_test\_accuracy}, \texttt{mean\_k5\_acc}, \texttt{tpp\_threshold\_10\_total\_metric}, \texttt{scr\_metric\_threshold\_10}, \texttt{autointerp\_score}, \texttt{disentangle\_score}.}
\label{tab:diagnostic_grid}
\centering
\small
\setlength{\tabcolsep}{4pt}
\begin{tabular}{lcccccc}
\toprule
SAE / Benchmark & sparse\_probing & sae-probes & TPP@10 & SCR@10 & autointerp & RAVEL disent. \\
\midrule
canonical                   & $0.858 \pm 0.004$ & $0.881 \pm 0.000$ & $0.046 \pm 0.029$ & $0.208 \pm 0.025$ & $0.790 \pm 0.010$ & $0.697 \pm 0.004$ \\
\texttt{permuted\_decoder}  & $0.857 \pm 0.005$ & $0.881 \pm 0.000$ & $0.004 \pm 0.008$ & $-0.002 \pm 0.039$ & $0.790 \pm 0.005$ & $0.290 \pm 0.002$ \\
\texttt{random\_l0\_matched}& $0.744 \pm 0.008$ & $0.745 \pm 0.003$ & $0.002 \pm 0.004$ & $-0.002 \pm 0.001$ & $0.694 \pm 0.012$ & $0.291 \pm 0.002$ \\
\bottomrule
\end{tabular}

\end{table}

\paragraph{Pass/fail summary.} \texttt{permuted\_decoder} \textbf{fails} sparse\_probing, sae-probes, and autointerp (each within reseed noise of canonical, indicating these benchmarks are encoder-only by construction); \textbf{passes} TPP, SCR, and RAVEL (each crashes well outside reseed noise). \texttt{random\_l0\_matched} \textbf{passes} all six benchmarks (each crashes well below canonical), but the no-information floor is non-trivially high on sparse\_probing ($0.744$) and sae-probes ($0.745$), so a casual benchmark consumer might mistake an L0-matched random projection for a weak but meaningful SAE.

\paragraph{Implementation notes.} SAEBench's RAVEL hooks assume \texttt{Gemma2DecoderLayer.forward} returns a \texttt{(hidden\_states,)} tuple; in transformers 5.6.2 it returns \texttt{hidden\_states} directly, breaking the hook code with \texttt{IndexError: too many indices for tensor of dimension 2}. Our runner monkey-patches the hook boundary; without this, all RAVEL runs fail. Worth flagging as a SAEBench compatibility bug. The \texttt{random\_l0\_matched} diagnostic uses a uniform threshold across all 16{,}384 latents (single value), whereas canonical Gemma-Scope JumpReLU uses a learned per-feature threshold (range $2.59$--$16.76$, mean $3.13$). The deviation is intentional (the random encoder has no learned feature structure that per-feature thresholds could exploit) but is worth flagging.

\section{Sanity check against public SAEBench baselines}
\label{app:hf_baselines}

To check that our SAEBench pipeline matches the official one, we fetched the per-SAE result JSONs published by the SAEBench authors on Hugging Face (\texttt{adamkarvonen/sae\_bench\_results}) for the canonical 16k Gemma Scope SAE (Gemma-2-2b layer 12, also used in Appendix~\ref{app:per_sae_noise}) and compared them to the means of our reseeded runs. Table~\ref{tab:hf_baselines} shows the side-by-side. Every metric matches the HF baseline to within $\pm 0.012$ except SCR top-500 ($-0.042$), well within our own multi-seed standard deviation in each case.

\begin{table}[h]
\caption{Side-by-side comparison of our reseeded means against the public SAEBench result JSONs for the canonical 16k Gemma Scope SAE. Sanity check that our pipeline reproduces the original work; differences are within reseed noise.}
\label{tab:hf_baselines}
\centering
\small
\begin{tabular}{lrrr}
\toprule
Benchmark & Ours (mean) & HF baseline & Ours $-$ baseline \\
\midrule
sparse\_probing top-1 acc & 0.756 & 0.760 & $-$0.003 \\
sparse\_probing top-5 acc & 0.871 & 0.878 & $-$0.007 \\
TPP top-2 & 0.009 & 0.009 & $+$0.001 \\
TPP top-10 & 0.037 & 0.025 & $+$0.012 \\
TPP top-50 & 0.138 & 0.140 & $-$0.002 \\
TPP top-500 & 0.400 & 0.394 & $+$0.005 \\
SCR top-10 & 0.287 & 0.284 & $+$0.004 \\
SCR top-500 & 0.302 & 0.344 & $-$0.042 \\
autointerp score & 0.820 & 0.828 & $-$0.008 \\
\bottomrule
\end{tabular}

\end{table}

\section{SCR task construction}
\label{app:scr_construction}

The SCR task generator iterates over four (T-structure, S-structure) combinations: T as a single in-sae feature or as the OR of two in-sae features, crossed with S in the same two structures. For each combination, the generator draws random in-sae feature pairs and keeps up to 3 that pass two filters: (i) T and S come from different L0 root subtrees of the SynthSAEBench-16k hierarchy (so neither sibling mutual exclusion nor parent-child gating prevents independent co-firing), and (ii) all four cells of the (T, S) firing contingency table have at least 100 samples on a 60{,}000-activation pool drawn for the seed. The maximum is therefore $4 \times 3 = 12$ tasks per seed; the actual count varies because random sampling occasionally fails to produce 3 viable pairs in 15 attempts (most often for the (OR, OR) combination, which has the strictest co-firing constraint). On our three task-generation seeds we obtained 9, 11, and 12 tasks respectively (Table~\ref{tab:scr_task_counts}).

We also attempted out-of-sae versions of these combinations (T or S as the OR of two out-of-sae features), but the (T=1 $\wedge$ S=1) cell never reached the 100-sample minimum: out-of-sae features fire too rarely on this model, and OR'ing two of them does not lift the joint rate enough. All retained SCR tasks are therefore in-sae.

\begin{table}[h]
\caption{Number of SCR tasks kept per seed, broken down by the structures of T and S. Each (T-structure, S-structure) combination targets 3 tasks but is capped at the number of random feature pairs (up to 15 draws) that satisfy both the different-root constraint and the $\geq 100$ samples per contingency cell minimum.}
\label{tab:scr_task_counts}
\centering
\small
\begin{tabular}{lccccc}
\toprule
seed & (single, single) & (single, or) & (or, single) & (or, or) & total \\
\midrule
1234 & 3 & 3 & 3 & 0 &  9 \\
2222 & 3 & 3 & 2 & 3 & 11 \\
3333 & 3 & 3 & 3 & 3 & 12 \\
\bottomrule
\end{tabular}
\end{table}

\section{SCR batch-size sweep on the v1 panel}
\label{app:scr_batch}

SAEBench flags \texttt{probe\_train\_batch\_size} as load-bearing for SCR. Table~\ref{tab:scr_batch} reports Spearman $\rho$ between SCR canonical (top-$n=10$) and GT-MCC on the 39-SAE v1 panel under a single task-generation seed (1234), as we sweep the batch size over $\{8, 16, 32, 64\}$ with all other hyperparameters at canonical defaults. The canonical batch size $16$ yields the highest $\rho$ against both ground-truth metrics; deviating to batch $64$ degrades the signal substantially. Against GT-F1 the picture is similar in shape, with batch $16$ also the best.

\begin{table}[h]
\caption{SCR canonical (top-$n=10$, \texttt{probe\_l1\_penalty}=$10^{-3}$) Spearman correlation with GT-MCC and GT-F1 across the 39-SAE v1 panel as a function of \texttt{probe\_train\_batch\_size}. Single task-generation seed (1234). The canonical batch size 16 maximizes both signals.}
\label{tab:scr_batch}
\centering
\small
\begin{tabular}{crr}
\toprule
\texttt{probe\_train\_batch\_size} & $\rho$ vs GT-MCC & $\rho$ vs GT-F1 \\
\midrule
8 & $+$0.72 & $+$0.65 \\
\textbf{16 (canonical)} & $\mathbf{+0.79}$ & $\mathbf{+0.70}$ \\
32 & $+$0.70 & $+$0.61 \\
64 & $+$0.54 & $+$0.49 \\
\bottomrule
\end{tabular}

\end{table}

\section{Cross-base-model variation results}
\label{app:variations}

To probe whether the correlations reported in \S\ref{sec:results} reflect properties of SynthSAEBench-16k v1 specifically rather than the benchmarks themselves, we re-run the synthetic-task pipeline on three base-model variations published alongside SynthSAEBench-16k:
\begin{itemize}
\item \texttt{rel-p-0.5}: per-feature firing probabilities are halved relative to v1, so the activation pool is sparser overall.
\item \texttt{rel-p-1.5}: per-feature firing probabilities are scaled up by $1.5\times$, so features fire more often.
\item \texttt{std-2.5}: the rectified-Gaussian magnitude distribution uses a $2.5\times$ larger standard deviation, so on-firing magnitudes vary more widely.
\end{itemize}
For each variation we train a fresh 5-SAE log-uniform-L0 panel from scratch on the variation's activations, then evaluate at the same canonical hyperparameters used in \S\ref{sec:results}. Each variation panel has 9 SAEs (5 trained + 3 degraded controls + 1 oracle) at a single task seed, so within-trained $\rho$ is sensitive to small-panel noise; rankings are stable within each variation, however, and the disagreement between rows reflects population-level $\rho$ on different feature distributions.

Table~\ref{tab:variations} gives the headline canonical numbers; Table~\ref{tab:variations_full} extends the comparison across the in / out / mixed task categories. Two findings emerge. First, SCR canonical's within-trained $\rho$ swings from $+0.90$ on \texttt{std-2.5} to $-0.10$ on \texttt{rel-p-1.5}, far beyond what task-sampling noise on a single underlying $\rho$ can explain, and the $-0.10$ value confirms that SCR can read negatively correlated with quality on a perfectly reasonable variant of the same base model. Second, sparse probing's calibration on out-of-sae features is itself firing-rate-dependent: single out-of-sae $\rho$ is $+0.13$ on \texttt{rel-p-0.5} (preserved), $+0.23$ on v1, and $+0.67$--$+0.73$ on the higher-firing-rate \texttt{rel-p-1.5} and \texttt{std-2.5} variations. When out-of-sae features fire often enough, hidden-space proxies for them become accessible to a probe via correlated in-sae latents, so the agnosticism we report on v1 is a property of v1's firing-rate regime rather than a guarantee. The $+0.00$ within-trained value for sparse probing on \texttt{rel-p-0.5} in Table~\ref{tab:variations} is dominated by a non-monotone GT-MCC-vs-L0 curve on that variation, not a calibration failure.

\begin{table}[h]
\caption{Cross-base-model behavior of three benchmarks at canonical hyperparameters with GT-MCC. Each cell is full-panel $\rho$ / within-trained-SAE $\rho$. SCR canonical's within-trained $\rho$ swings from $+0.90$ to $-0.10$ across base models, beyond what task-sampling noise on a single underlying $\rho$ can explain. The $+0.00$ on \texttt{rel-p-0.5} for sparse probing is dominated by a non-monotone GT-MCC-vs-L0 curve on that variation rather than a calibration failure.}
\label{tab:variations}
\centering
\small
\setlength{\tabcolsep}{4pt}
\begin{tabular}{lcccc}
\toprule
Benchmark & v1 & \texttt{std-2.5} & \texttt{rel-p-0.5} & \texttt{rel-p-1.5} \\
 & (3 seeds) & (1 seed) & (1 seed) & (1 seed) \\
\midrule
Sparse probing single in-sae (top-16) & $+0.58 / +0.49$ & $+0.98 / +0.90$ & $+0.68 / +0.00$ & $+1.00 / +1.00$ \\
TPP all out-of-sae (top-10) & $+0.38 / +0.42$ & $-0.03 / +0.90$ & $+0.17 / +0.60$ & $+0.18 / +1.00$ \\
\textbf{SCR canonical (top-10)} & $\mathbf{+0.64 / +0.65}$ & $\mathbf{+0.64 / +0.90}$ & $\mathbf{+0.55 / +0.30}$ & $\mathbf{+0.13 / -0.10}$ \\
\bottomrule
\end{tabular}

\end{table}

\begin{table}[h]
\caption{Full-panel Spearman $\rho$ between each benchmark category and GT-MCC across the four base-model variations, at canonical hyperparameters (top-$k=16$ for sparse probing; top-$n=10$ for TPP and SCR). Sparse probing single in-sae stays positive across all four variations, but single out-of-sae $\rho$ rises sharply on the higher-firing-rate variations (\texttt{rel-p-1.5}, \texttt{std-2.5}), indicating that sparse probing's silence on out-of-sae features depends on the base model's firing-rate regime. SCR canonical is positive on three variations and weakly positive on \texttt{rel-p-1.5}, but its within-trained $\rho$ on \texttt{rel-p-1.5} (Table~\ref{tab:variations}) is the headline failure.}
\label{tab:variations_full}
\centering
\small
\setlength{\tabcolsep}{4pt}
\begin{tabular}{lcccc}
\toprule
Benchmark category & v1 & \texttt{std-2.5} & \texttt{rel-p-0.5} & \texttt{rel-p-1.5} \\
\midrule
Sparse probing single in-sae   & $+0.58$ & $+0.98$ & $+0.68$ & $+1.00$ \\
Sparse probing single out-of-sae & $+0.23$ & $+0.67$ & $+0.13$ & $+0.73$ \\
TPP all in-sae                 & $+0.18$ & $+0.75$ & $+0.65$ & $+0.77$ \\
TPP all out-of-sae             & $+0.38$ & $-0.03$ & $+0.17$ & $+0.18$ \\
SCR canonical (T=in, S=in)     & $+0.64$ & $+0.64$ & $+0.55$ & $+0.13$ \\
\bottomrule
\end{tabular}

\end{table}

\section{Ground-truth F1 and MCC vs L0 per SAE architecture}
\label{app:gt_vs_l0}

\begin{figure}[h]
\centering
\includegraphics[width=0.85\linewidth]{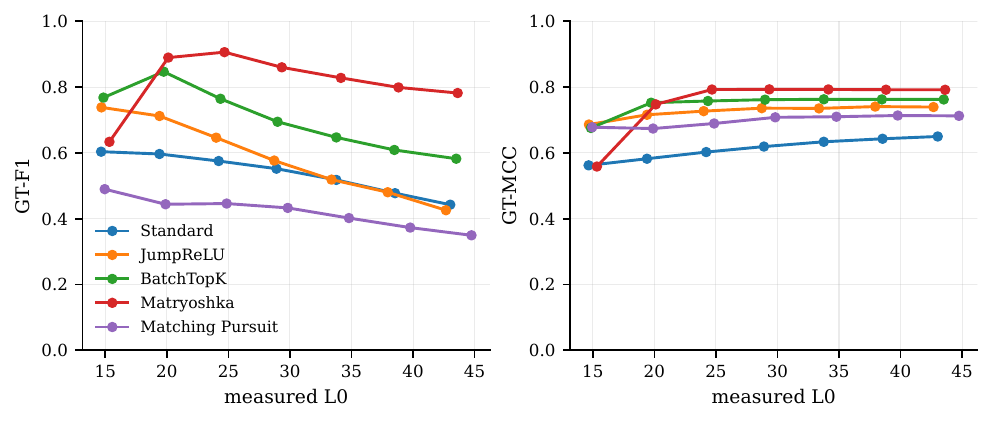}
\caption{GT-F1 (left) and GT-MCC (right) vs measured L0 for each SAE architecture in the v1 panel. Each point is one trained SAE (mean across task-generation seeds). GT-F1 has a clear optimum or monotone decay; GT-MCC stays high or rises gently across the same L0 range.}
\label{fig:gt_vs_l0}
\end{figure}

Figure~\ref{fig:gt_vs_l0} plots GT-F1 and GT-MCC as a function of measured L0 for each of the five SAE architectures in the v1 panel (Standard, JumpReLU, BatchTopK, Matryoshka, Matching Pursuit), at the seven trained L0 targets ($\{15, 20, 25, 30, 35, 40, 45\}$). GT-F1 declines monotonically with L0 on Standard, JumpReLU, BatchTopK, and Matching Pursuit, peaks around L0$\approx 25$--$30$ on Matryoshka, and is everywhere noticeably below 1 even on the best-performing variants. GT-MCC, in contrast, either rises gently with L0 or plateaus, and stays relatively high (above 0.55) across the full L0 range. This is a property of the two metrics rather than of the SAEs: GT-F1 penalizes over-firing latents (any latent that fires when its target ground-truth feature is off lowers precision), so increasing L0 past the underlying feature firing rate degrades F1; GT-MCC tolerates over-firing as long as the encoder still discriminates target-feature presence above noise, so it is much less sensitive to the L0-target axis. Because the proxy benchmarks studied in this paper consume only the encoder's output, sparse probing's better correlation with GT-MCC than GT-F1 (Table~\ref{tab:realsae_correlations}) is consistent with the proxy and ground-truth metric agreeing on what they tolerate.

\section{Hyperparameter sweep for sparse probing, TPP, and SCR}
\label{app:hparam_sweep}

\begin{figure}[h]
\centering
\includegraphics[width=0.65\linewidth]{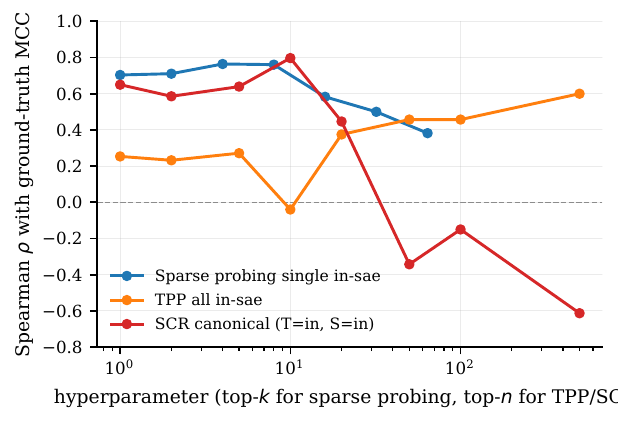}
\caption{Spearman $\rho$ between benchmark score and GT-MCC across the v1 synthetic panel as a function of the benchmark's primary hyperparameter (top-$k$ for sparse probing, top-$n$ for TPP and SCR), single seed (1234). Canonical-hparam points (top-$k=16$ for sparse probing, top-$n \in \{10, 50, 500\}$ for SCR, top-$n=10$ for TPP) use the full 35-trained-SAE panel; non-canonical points use the 15-trained-SAE sub-panel from the original L0 set $\{15, 25, 45\}$. Sparse probing stays positive across the full range of $k$. SCR's correlation flips from positive at top-$n \leq 20$ to strongly negative at top-$n=500$ ($\rho = -0.61$ on the 35-SAE panel). TPP's correlation in-sae stays in the $-0.04$ to $+0.60$ range and dips to near zero at canonical top-$n=10$.}
\label{fig:hparam_sweep}
\end{figure}

We calculate Spearman $\rho$ between benchmark score and GT-MCC across the v1 synthetic panel as a function of the benchmark's primary hyperparameter (top-$k$ for sparse probing, top-$n$ for TPP and SCR), and show results in Figure~\ref{fig:hparam_sweep}.

\section{Per-task / per-hyperparameter validity scatter grids}
\label{app:validity_grids}

\begin{figure}[h]
\centering
\includegraphics[width=\linewidth]{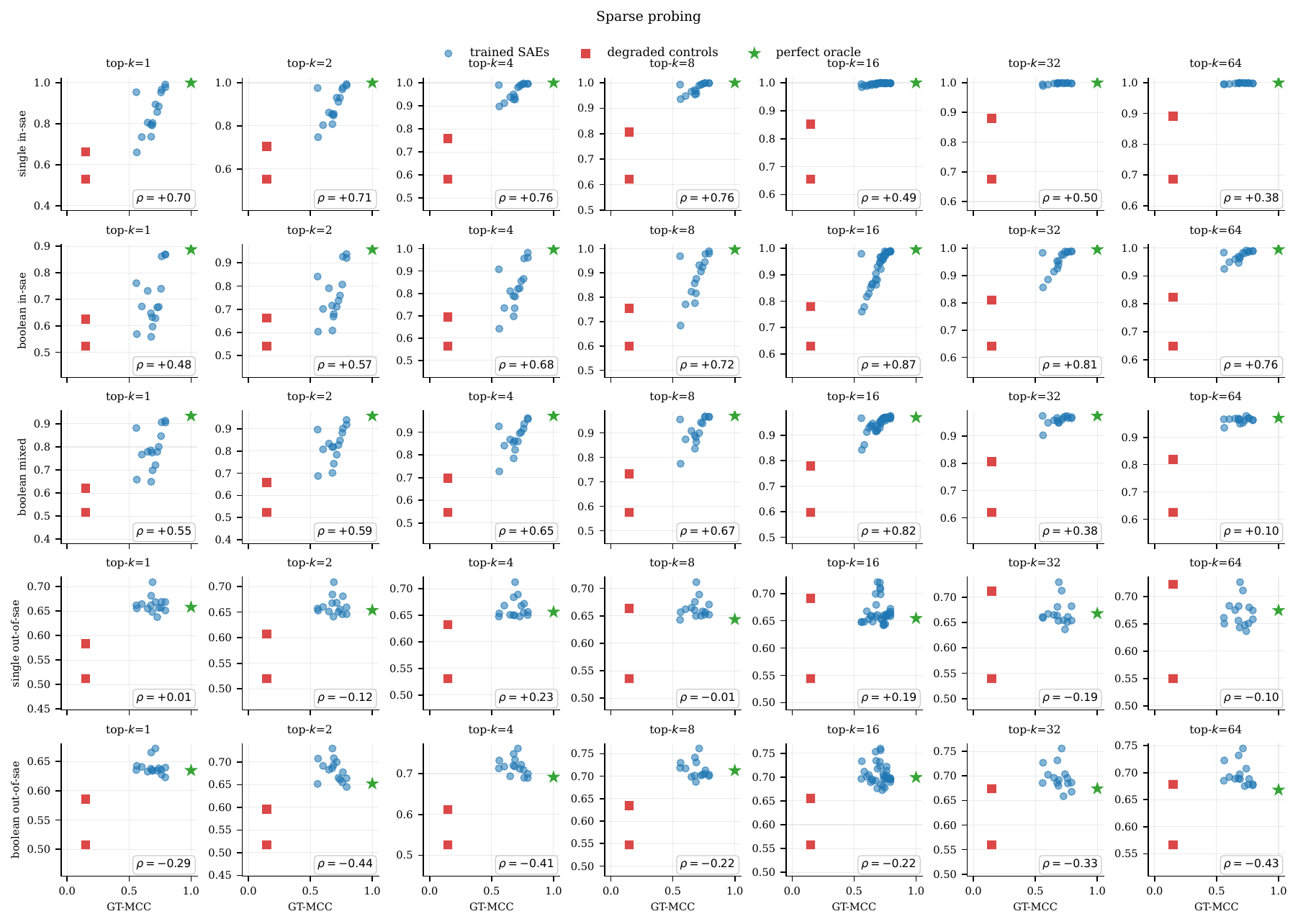}
\caption{Sparse probing: GT-MCC vs benchmark score across all task categories (rows) and top-$k$ values (columns). Decoder-only degraded controls (\texttt{permuted\_decoder}) are excluded because sparse probing is encoder-only and would score them identically to the canonical SAE. The boolean-mixed row and the $k=16$ column correspond to the panel shown in Figure~\ref{fig:validity_scatter}.}
\label{fig:validity_grid_sp}
\end{figure}

\begin{figure}[h]
\centering
\includegraphics[width=\linewidth]{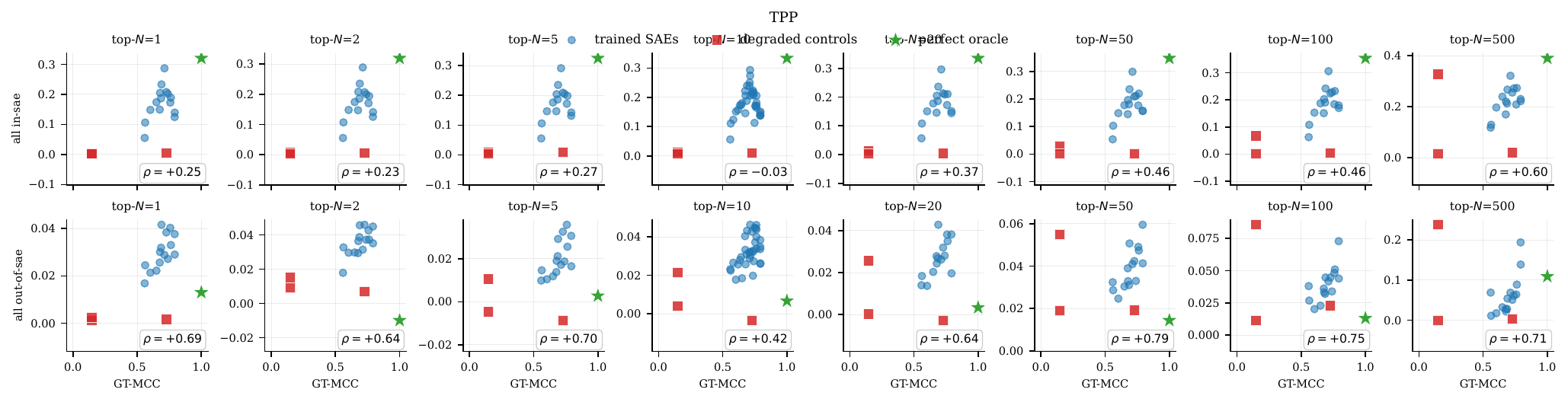}
\caption{TPP: GT-MCC vs benchmark score across all-in-sae and all-out-of-sae sibling groups (rows) and top-$n$ values (columns). The all-in-sae row at $n=10$ corresponds to the panel shown in Figure~\ref{fig:validity_scatter}; out-of-sae correlations are higher than in-sae correlations across the entire $n$ range, the ``leak'' discussed in \S\ref{sec:results}.}
\label{fig:validity_grid_tpp}
\end{figure}

\begin{figure}[h]
\centering
\includegraphics[width=\linewidth]{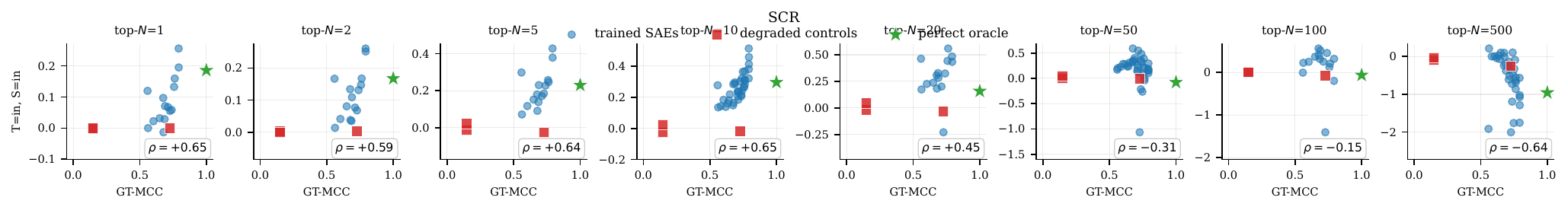}
\caption{SCR (T=in, S=in): GT-MCC vs benchmark score across top-$n$ values. The $n=10$ panel corresponds to Figure~\ref{fig:validity_scatter}; correlations swing from positive at small $n$ to strongly negative by $n=500$.}
\label{fig:validity_grid_scr}
\end{figure}

Figure~\ref{fig:validity_scatter} in \S\ref{sec:gt} shows benchmark score vs GT-MCC at canonical hyperparameters for one task category each. For completeness, this appendix gives the full ($\mathrm{category} \times \mathrm{hparam}$) grid of scatter panels for sparse probing (Figure~\ref{fig:validity_grid_sp}), TPP (Figure~\ref{fig:validity_grid_tpp}), and SCR (Figure~\ref{fig:validity_grid_scr}). Single task-generation seed (1234); each panel reports its multi-seed mean Spearman $\rho$ where multi-seed data exists, and its single-seed Spearman otherwise. Canonical-hparam columns (top-$k=16$ for sparse probing, top-$n=10$ for TPP, and top-$n \in \{10, 50, 500\}$ for SCR) use the full 35-trained-SAE panel; the other columns use the 15-trained-SAE sub-panel.

\section{Practitioner summary across the suite}
\label{app:summary}

We provide a Practitioner summary across the SAEBench metrics audited in this study in Table~\ref{tab:summary}.

\begin{table}[h]
\caption{Practitioner summary across the SAEBench metrics audited in this study. Reseed CV is from the canonical 65k Gemma Scope SAE (Table~\ref{tab:noise}); see Appendix~\ref{app:per_sae_noise} for the 16k SAE. Validity $\rho$ is the multi-seed mean Spearman correlation with GT-MCC across the 35-trained-SAE v1 panel (degraded controls and the oracle excluded). SNR is the ANOVA-style between/within ratio on the 4-SAE architecture training panel (see Appendix~\ref{app:snr_full}, Figure~\ref{fig:snr_bars_full}); ranges span hyperparameter sweeps within the metric.}
\label{tab:summary}
\centering
\small
\setlength{\tabcolsep}{4pt}
\begin{tabular}{lcccl}
\toprule
Metric & Reseed CV & Validity $\rho$ & SNR & Direction with training \\
       & (\S\ref{sec:noise}) & (\S\ref{sec:gt}) & (\S\ref{sec:snr}) & \\
\midrule
\multicolumn{5}{l}{\textit{Report by default.}} \\
core (MSE / EV / KL / CE)         & --      & --       & 8--14    & monotone improving \\
sae-probes ($k$=1\dots16)         & 0.2\%   & high     & 5--7     & monotone improving \\
sparse probing (top-$k$)          & 0.3--1.2\%   & $+0.49$ to $+0.87$  & 1.8--4.2 & monotone improving \\
RAVEL disentangle                 & 0.2\%   & --       & 2.6      & monotone improving \\
autointerp ($n_{\text{lat}}$=1k)  & 0.5\%   & --       & 1.8      & monotone improving \\
\midrule
\multicolumn{5}{l}{\textit{Avoid: sign-flips, low SNR, or declines with training.}} \\
SCR                               & 3--7\%  & $-0.64$ to $+0.65$ & 1.5--3.9 & declines on 3/4 variants \\
TPP                               & 3--39\% & $\approx 0$        & 1.1--3.9 & improves low $N$, declines high $N$ \\
RAVEL cause / isolation           & 2--3\%  & --       & 0.7--1.0 & coin-flip on snapshots \\
\bottomrule
\end{tabular}

\end{table}

\section{SAE training setup for the discriminability panels}
\label{app:training_setup}

The two panels of trained SAEs analyzed in \S\ref{sec:snr} (cross-architecture and log-uniform-prefix Matryoshka) share most training hyperparameters and differ only in architecture-specific configuration. We document them here to support reproducibility and to explain the log-uniform-prefix Matryoshka variant, which is non-standard.

\paragraph{Common settings.} All SAEs are trained on the residual stream after layer~12 of \texttt{google/gemma-2-2b} (\texttt{blocks.12.hook\_resid\_post}), $d_{\text{in}}=2304$, on The Pile~\citep{pile} with context length 1024. Every SAE has $d_{\text{sae}} = 32{,}768$ ($16\times d_{\text{in}}$). We use Adam with $\beta_1=0.9$ (SAELens defaults: $\beta_2=0.999$, $\epsilon=10^{-8}$, weight decay $0$), peak learning rate $3\times10^{-4}$, no warm-up, and a linear LR decay over the final fifth of training. Training batch size is 4{,}096 tokens. Activations are accessed through \texttt{transformer\_lens} with \texttt{exclude\_special\_tokens=True} and \texttt{prepend\_bos=True}. Mixed-precision (bf16 autocast) is enabled for both the language model and the SAE. The cross-architecture panel uses seed 0 for every variant; the sampled-Matryoshka panel uses seeds $\{0, 1, 2\}$ for every variant (Appendix~\ref{app:multi_seed}).

\paragraph{Cross-architecture panel (1.5B tokens).} Four SAEs trained for $1.5\times10^{9}$ tokens each (366{,}210 steps; LR decays over the final $3\times10^{8}$ tokens). The four variants differ only in the sparsifying activation:
\begin{itemize}\setlength{\itemsep}{0pt}
\item \textbf{BatchTopK} ($k\in\{50, 100\}$), one decoder; the global top-$k\cdot B$ activations across the batch are kept active per step \citep{bussmann2024batchtopk}.
\item \textbf{Matryoshka BatchTopK} ($k\in\{50, 100\}$) with three nested inner widths $\{2{,}048,\,8{,}192,\,32{,}768\}$ (i.e.\ $d_{\text{sae}}/16$, $d_{\text{sae}}/4$, $d_{\text{sae}}$). At each step the loss is the sum of BatchTopK reconstruction losses over the three prefixes (and the auxiliary dead-feature term), so each prefix learns to be a self-contained dictionary \citep{bussmann2025learning}. Inner-width selection is fixed across training steps.
\end{itemize}
We snapshot 28 times per SAE (token counts $\{0, 10\text{M}, 25\text{M}, 50\text{M}, 100\text{M}, 150\text{M}, \ldots, 1{.}0\text{B}, 1{.}1\text{B}, \ldots, 1{.}5\text{B}\}$). The token-0 (random-init) snapshot is dropped from all training-trajectory analyses.

\paragraph{Log-uniform-prefix Matryoshka panel (300M tokens).} Four Matryoshka BatchTopK SAEs ($k=100$) per training seed, trained for $3\times10^{8}$ tokens each (73{,}242 steps; LR decays over the final $6\times10^{7}$ tokens). Variants differ only in $n \in \{1, 2, 3, 4\}$, the number of inner-width prefixes sampled per training step. We train each variant from three independent random seeds (0, 1, 2), giving $4 \times 3 = 12$ SAE training runs in total (Appendix~\ref{app:multi_seed}). Each (variant, seed) takes 11 snapshots at $\{0, 30\text{M}, 60\text{M}, \ldots, 300\text{M}\}$ tokens.

The non-standard part is the inner-width selection rule. Rather than a fixed prefix list, at each training step we draw $n$ widths from a log-uniform distribution and use them as that step's matryoshka prefixes (always including the full $d_{\text{sae}}$ as the outermost prefix):
\[
\log(w_i) \sim \mathrm{Uniform}\!\left(\log w_{\min},\ \log w_{\max}\right), \quad w_{\min}=64, \quad w_{\max}=d_{\text{sae}}=32{,}768.
\]
The $n$ samples are deduplicated, sorted, and clamped to $[w_{\min}, w_{\max}-1]$, and the Matryoshka loss is computed against the resulting prefixes for that step only. Different steps see different prefix sets, so the SAE is trained as a Matryoshka over the entire log-spaced range $[64, 32{,}768]$ rather than over a small fixed prefix list. A higher $n$ trades wall-clock speed for variance reduction in the per-step prefix coverage. To our knowledge the same log-uniform prefix-sampling scheme is used by Gemma Scope 2's Matryoshka SAEs~\citep{mcdougall2025gemmascope2}; we adopt it here to mirror that setup.

The intuition behind log-uniform sampling is that ``how good is this SAE at a given prefix width $w$?'' is most naturally a question about $\log w$ rather than $w$: doubling $w$ from 64 to 128 changes capacity as much as doubling from 16{,}384 to 32{,}768. Log-uniform sampling spreads training signal evenly across that scale rather than concentrating it on widths near $d_{\text{sae}}$, which a uniform-in-$w$ rule would do.

\section{Compute resources}
\label{app:compute}

All experiments were run on a single NVIDIA H100 (80GB) GPU at a time; we did not parallelize a single experiment across multiple GPUs. Wall-clock numbers below are approximate, since (a) several experiments were re-run for the L0 fill-in and the reseed sweep, and (b) we share H100s with other users.

\paragraph{LLM SAE training (Section~\ref{sec:snr}).}
\begin{itemize}\setlength{\itemsep}{0pt}
\item \emph{Cross-architecture panel}: each of the 4 SAEs (1.5B tokens) takes $\sim$12~hours on an H100 $\Rightarrow$ $\sim$48 H100-hours for the panel (single training seed).
\item \emph{Sampled-Matryoshka panel}: each of the $4 \times 3 = 12$ SAEs (4 variants $\times$ 3 training seeds, 300M tokens each) takes $\sim$3~hours on an H100 $\Rightarrow$ $\sim$36 H100-hours for the panel.
\end{itemize}

\paragraph{Synthetic-data SAE training (Section~\ref{sec:gt}).} The 35 SAEs trained on SynthSAEBench-16k (5 architectures $\times$ 7 L0 levels, 200M synthetic samples each, $d_{\text{sae}}=4096$) are much faster than the LLM SAEs because there is no language-model forward pass. Each takes roughly 20--30 minutes on an H100; the full panel is well under 20 H100-hours.

\paragraph{SAEBench evaluations (Sections~\ref{sec:noise}, \ref{sec:snr}).} Per-SAE wall clock varies widely across evaluations: sparse probing, TPP, SCR, and RAVEL are each 20--40 minutes on an H100; autointerp is mostly bound by OpenAI API latency rather than H100 time; core and sae-probes are 5--15 minutes once the LM activations are cached. The reseed-noise study (\S\ref{sec:noise}) covers 4 canonical SAEs $\times$ 5 reseeds $\times$ 6 evaluations and totals $\sim$40 H100-hours plus $\sim$\$30 of OpenAI API spend on autointerp. The snapshot evaluations dominate total compute: 28 snapshots per cross-architecture SAE and 11 snapshots per sampled-Matryoshka SAE, $\times$ 6 evaluations across the $4 + 12 = 16$ trained SAEs (4 cross-architecture and 12 sampled-Matryoshka after multi-seed expansion), comes to $\sim$500--650 H100-hours.

\paragraph{Synthetic-validity evaluations (Section~\ref{sec:gt}).} Each of the 40 SAEs in the synthetic panel (across the v1 and three SAE-variation panels) is encoded once on a 60{,}000-activation pool and then run through 92 sparse-probing tasks, 60 TPP sibling groups, and up to 12 SCR pairs per task-generation seed. Per SAE this is well under an hour on an H100; across 3 task seeds for v1 and a single seed for each variation, the panel costs $\sim$60 H100-hours total.

\paragraph{Preliminary and exploratory experiments.} The numbers above are for the experiments reported in this paper. The full project also included exploratory runs (alternative L0 schedules, alternative reseed-noise SAEs, batch-size and probe-regularizer sweeps, hyperparameter ablations) that we did not include because they did not change the qualitative findings. We estimate the total project compute at roughly 2--3$\times$ the above, dominated again by SAEBench snapshot evaluations on intermediate SAE checkpoints.

\section{Probe-ablation training-trajectory slopes}
\label{app:trajectories}

Table~\ref{tab:slopes} reports linear-regression slopes of TPP and SCR scores against training tokens (in score units per billion tokens, after a 100M-token warmup) for every threshold across the four-SAE training panel. Positive values indicate improvement through training; negative values indicate decline. TPP shows a clean cross-over between top-$20$ and top-$50$ (positive slopes for all four variants below the cross-over, negative for all four above). SCR slopes are negative across most thresholds for both BTK variants and at top-$N \geq 50$ for the Matryoshka variants.

\begin{table}[h]
\caption{Slopes of TPP and SCR scores against training tokens (score per B tokens after a 100M warmup) on the four-SAE training panel. TPP shows a clean cross-over between top-$20$ and top-$50$: positive (improving) for all four variants at top-$N \leq 10$ and at top-$N=20$ for three of four, then negative (declining) for all four at top-$N \geq 50$. SCR is negative across most thresholds for the BTK variants and at top-$N \geq 50$ for the Matryoshka variants.}
\label{tab:slopes}
\centering
\small
\setlength{\tabcolsep}{4pt}
\begin{tabular}{lrrrrrrr}
\toprule
\textbf{TPP} top-$N$ & $2$ & $5$ & $10$ & $20$ & $50$ & $100$ & $500$ \\
\midrule
BTK $k$=50         & $+0.039$ & $+0.057$ & $+0.010$ & $-0.017$ & $-0.048$ & $-0.046$ & $-0.030$ \\
BTK $k$=100        & $+0.016$ & $+0.048$ & $+0.028$ & $+0.014$ & $-0.029$ & $-0.056$ & $-0.062$ \\
Matryoshka $k$=50  & $+0.037$ & $+0.025$ & $+0.025$ & $+0.002$ & $-0.021$ & $-0.028$ & $-0.026$ \\
Matryoshka $k$=100 & $+0.019$ & $+0.054$ & $+0.039$ & $+0.007$ & $-0.021$ & $-0.029$ & $-0.046$ \\
\midrule
\textbf{SCR} top-$N$ & $2$ & $5$ & $10$ & $20$ & $50$ & $100$ & $500$ \\
\midrule
BTK $k$=50         & $-0.040$ & $-0.028$ & $-0.065$ & $-0.065$ & $-0.077$ & $-0.083$ & $-0.038$ \\
BTK $k$=100        & $-0.019$ & $-0.030$ & $-0.033$ & $-0.070$ & $-0.066$ & $-0.103$ & $-0.060$ \\
Matryoshka $k$=50  & $-0.009$ & $+0.022$ & $+0.005$ & $+0.008$ & $-0.025$ & $-0.046$ & $-0.066$ \\
Matryoshka $k$=100 & $+0.003$ & $-0.003$ & $-0.006$ & $+0.009$ & $-0.016$ & $+0.047$ & $-0.004$ \\
\bottomrule
\end{tabular}

\end{table}

\section{SNR ranking of SAEBench metrics on the training panels}
\label{app:snr_full}

\paragraph{Definition.} We summarize each metric by an ANOVA-style signal-to-noise ratio: the standard deviation across variants of each variant's mean score after a 100M-token warm-up (between-variant variation), divided by the pooled standard deviation of detrended within-variant residuals (within-variant noise, with a degree-2 polynomial fit removing monotone trajectory motion). $\mathrm{SNR} = 1$ means a metric's spread across variants matches its own per-checkpoint jitter; $\mathrm{SNR} \geq 2$ means a metric reads a comfortably stable inter-variant signal.

\begin{figure}[h]
\centering
\includegraphics[width=0.78\linewidth]{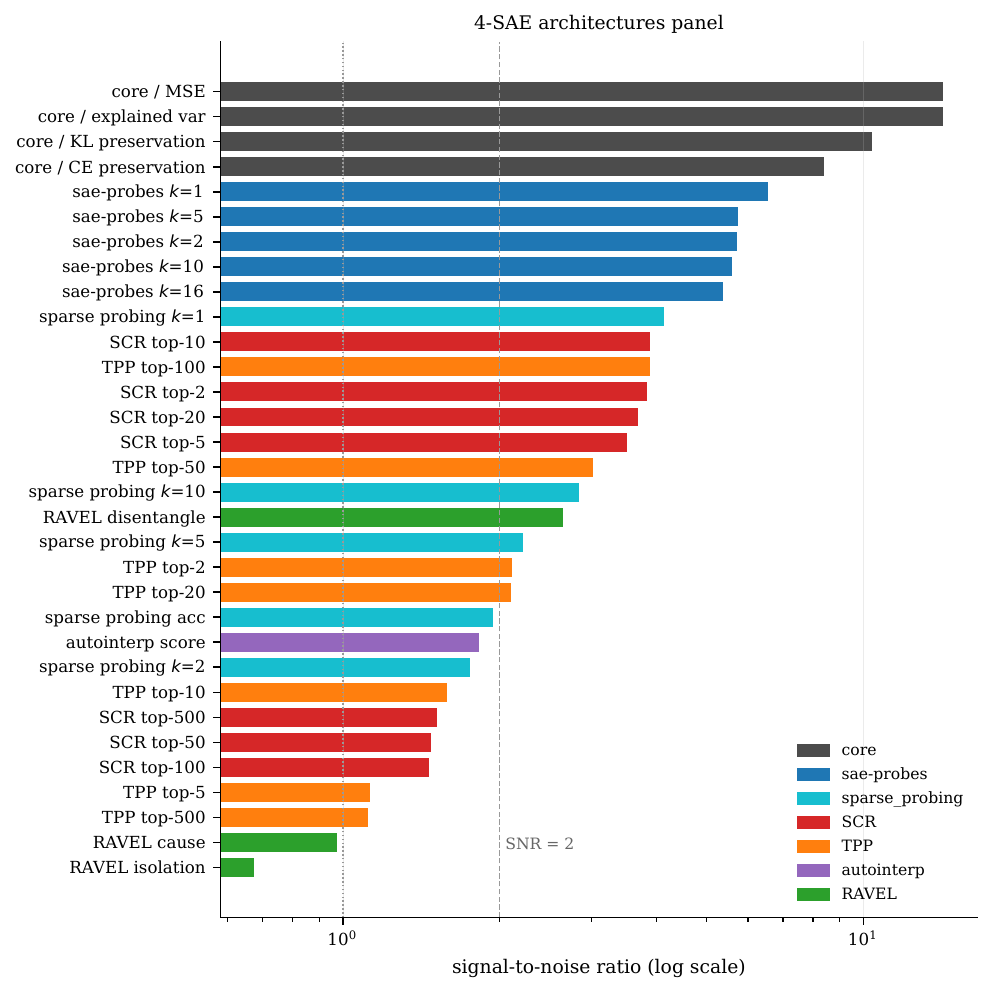}
\caption{Signal-to-noise ratio of every SAEBench metric on the four-SAE architecture snapshot panel, ordered by SNR (log scale). Reference dashed line at SNR $=2$. The Matryoshka-panel analogue is the variance decomposition in Figure~\ref{fig:multi_seed_vd}.}
\label{fig:snr_bars_full}
\end{figure}

\paragraph{Results.} Figure~\ref{fig:snr_bars_full} shows the SNR ranking for every SAEBench metric on the four-SAE architecture panel. Core reconstruction metrics dominate (MSE and explained variance both at SNR $14.2$, KL preservation $10.4$, CE preservation $8.4$). Among non-core metrics, sae-probes is the most discriminative (SNR $5.4$--$6.6$ across $k = 1\ldots16$). Default SAEBench sparse probing is roughly two-thirds as discriminative (top-$1$ SNR $4.2$, degrading to $2.8$ at top-$10$). SCR and TPP read modest signals at moderate thresholds (SCR top-$\{2,5,10,20\}$: $3.5$--$3.9$; TPP top-$\{50, 100\}$: $3.0$--$3.9$), while several published metrics have SNR $\leq 2$: RAVEL cause/isolation ($1.0$--$0.7$), autointerp ($1.8$), TPP top-$\{5, 10, 500\}$ ($1.1$--$1.6$), SCR top-$\{50, 100, 500\}$ ($1.5$). For these latter metrics inter-variant spread is comparable to per-checkpoint jitter, so picking a winner between SAEs in this panel from one checkpoint is roughly a coin flip.

We do not show an analogous SNR ranking for the sampled-Matryoshka panel: with three training seeds available there, the variance decomposition in Appendix~\ref{app:multi_seed} (Figure~\ref{fig:multi_seed_vd}) cleanly separates between-variant signal from training-seed and snapshot noise, which a single-number SNR summary cannot. The qualitative ordering on the Matryoshka panel is broadly consistent with the architectures panel (core reconstruction at the top; sae-probes and RAVEL disentangle next; SCR and TPP at the bottom), though variance shares are much smaller because the four Matryoshka SAEs are closer in absolute score than the four arch-panel SAEs. We do still compute the same SNR statistic on the Matryoshka panel and use it as the SNR-$\geq\!1$ filter for the cross-metric ranking heatmap (Appendix~\ref{app:contradiction}, Figure~\ref{fig:contradiction}); we just do not plot the resulting ordering separately because Figure~\ref{fig:multi_seed_vd} subsumes it.

\section{Cross-metric ranking agreement (contradiction matrices)}
\label{app:contradiction}

For each panel we compute pairwise Spearman $\rho$ of metric rankings using each metric's mean trajectory value per variant (post-warmup), oriented so higher $=$ better SAE. We restrict to SNR-informative metrics (SNR $\geq 1$, see Appendix~\ref{app:snr_full}). Figure~\ref{fig:contradiction} shows a representative subset of nine metrics on each panel; the title of each subplot reports the mean off-diagonal $\rho$ across the full SNR-informative metric set on that panel.

The cross-architecture panel (left) shows broad agreement (mean $\rho = +0.44$ across 30 informative metrics): when SAEs are very different from each other, most SAEBench metrics rank them similarly. The Matryoshka panel (right) shows essentially no agreement (mean $\rho = +0.08$ across 17 informative metrics) and several stark contradictions, including default sparse probing $k$=1 vs $k$=10 negatively correlated at $\rho = -0.4$ (a single metric family disagreeing with itself), and SCR top-$100$ disagreeing with core MSE, RAVEL disentangle, and SCR top-$\{2,10,500\}$ at $\rho = -0.2$ to $-0.4$. The two panels combined: SAEBench reliably ranks SAEs only when they are obviously different.

\begin{figure}[h]
\centering
\includegraphics[width=\linewidth]{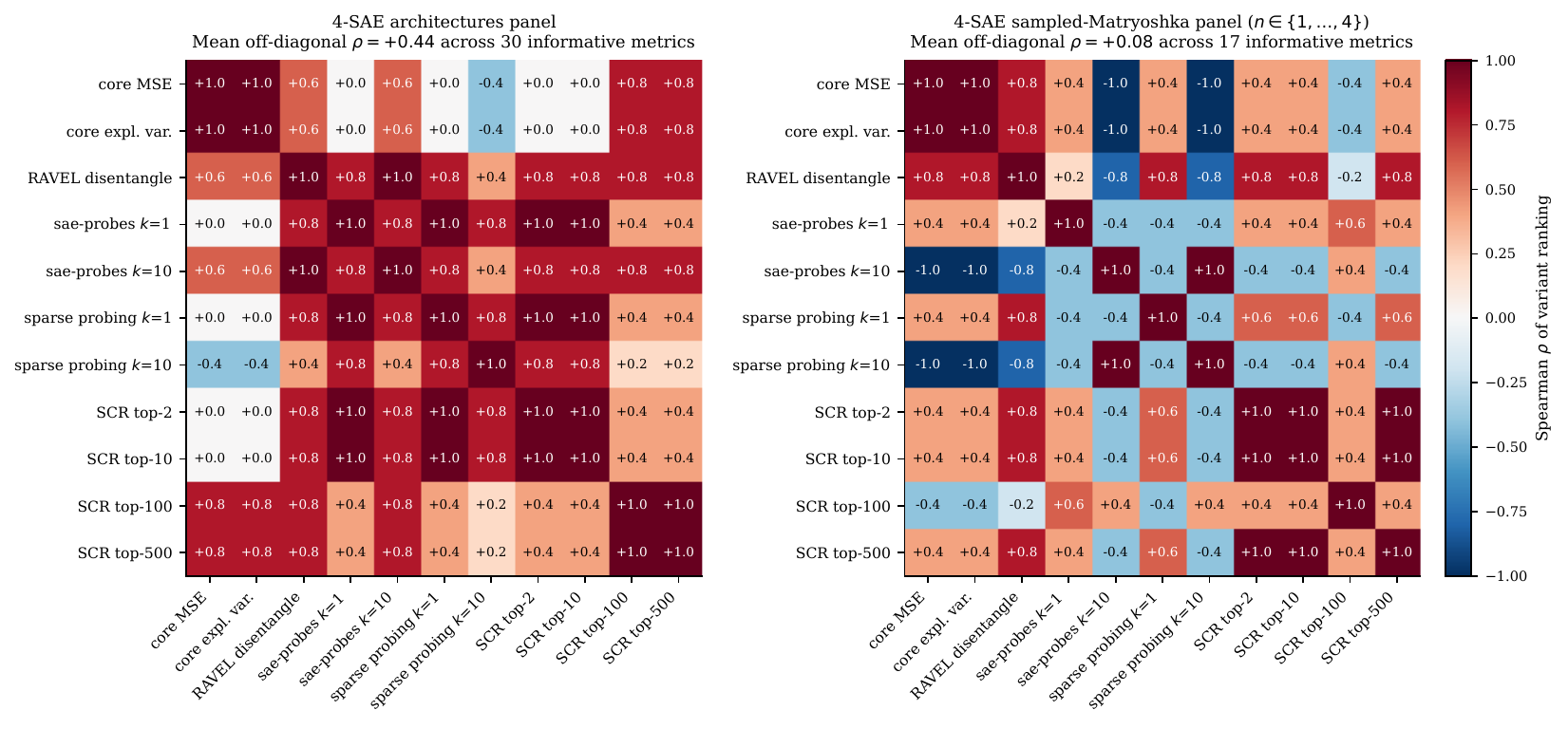}
\caption{Pairwise Spearman $\rho$ between SAEBench metric rankings on the four-SAE architectures panel (left) and the four-SAE sampled-Matryoshka panel (right; $n \in \{1,2,3,4\}$). Each cell is the rank correlation between two metrics' final-snapshot rankings of the four SAEs in that panel, oriented so positive $\rho$ means agreement on which SAE is better. Subtitle reports the mean off-diagonal $\rho$ across all SNR-informative metrics (SNR $\geq 1$); the displayed subset of nine metrics is representative. The Matryoshka panel uses seed 0; the multi-seed analysis is in Appendix~\ref{app:multi_seed}.}
\label{fig:contradiction}
\end{figure}

\section{Multi-seed variance decomposition on the Matryoshka panel}
\label{app:multi_seed}

This appendix gives the multi-seed analysis of the four-variant sampled-Matryoshka panel from \S\ref{sec:snr}. Each variant ($n \in \{1, 2, 3, 4\}$) was trained from three independent random seeds (0, 1, 2), all other settings as in Appendix~\ref{app:training_setup}. After dropping the token-0 random-init snapshot and applying a 60M-token warm-up, each (variant, seed) trajectory contributes 9 snapshot scores per metric, for a total of $4 \times 3 \times 9 = 108$ post-warmup observations per metric. Throughout this appendix ``audited metrics'' refers to the 34 SAEBench scores we report on this panel.

\subsection{Variance decomposition}
\label{app:multi_seed:vd}

\paragraph{Definitions.} For each metric, let $x_{v, s, t}$ denote the post-warmup score at variant $v$, seed $s$, snapshot $t$. We compute three variance components,
\begin{align*}
V_\text{variant}        &= \mathrm{Var}_v\!\Big(\mathrm{mean}_{s,t}\, x_{v, s, t}\Big), \\
V_\text{seed | variant} &= \mathrm{mean}_v\!\Big(\mathrm{Var}_s\!\big(\mathrm{mean}_t\, x_{v, s, t}\big)\Big), \\
V_\text{snap}           &= \mathrm{mean}_{v, s}\!\Big(\mathrm{Var}_t\, x_{v, s, t}\Big),
\end{align*}
i.e.\ the variance of variant means (signal), the average across variants of variance across seeds within a variant (training-seed noise), and the average within-trajectory variance (snapshot wiggle). We report each as a share of the sum, e.g.\ $\mathrm{share}_\text{variant} = V_\text{variant} / (V_\text{variant} + V_\text{seed | variant} + V_\text{snap})$. This share has a similar functional form to a one-way random-effects intraclass correlation coefficient but is not equivalent; we report it as a variance share.

\paragraph{Null calibration.} Under i.i.d.\ standard-normal noise on the same panel design (4 variants $\times$ 3 seeds $\times$ 9 snapshots), $\mathrm{share}_\text{variant}$ is bounded above zero by chance alone. A 5{,}000-trial Monte-Carlo simulation gives Pr$(\mathrm{share}_\text{variant} > 0.20) < 10^{-3}$, Pr$(\mathrm{share}_\text{variant} > 0.10) \approx 0.02$, and Pr$(\mathrm{share}_\text{variant} > 0.05) \approx 0.21$. We treat $\mathrm{share}_\text{variant} > 0.20$ as reliable evidence of inter-variant signal, $[0.10, 0.20]$ as weak evidence, and $< 0.10$ as consistent with no detectable signal on this panel.

\begin{figure}[h]
\centering
\includegraphics[width=0.78\linewidth]{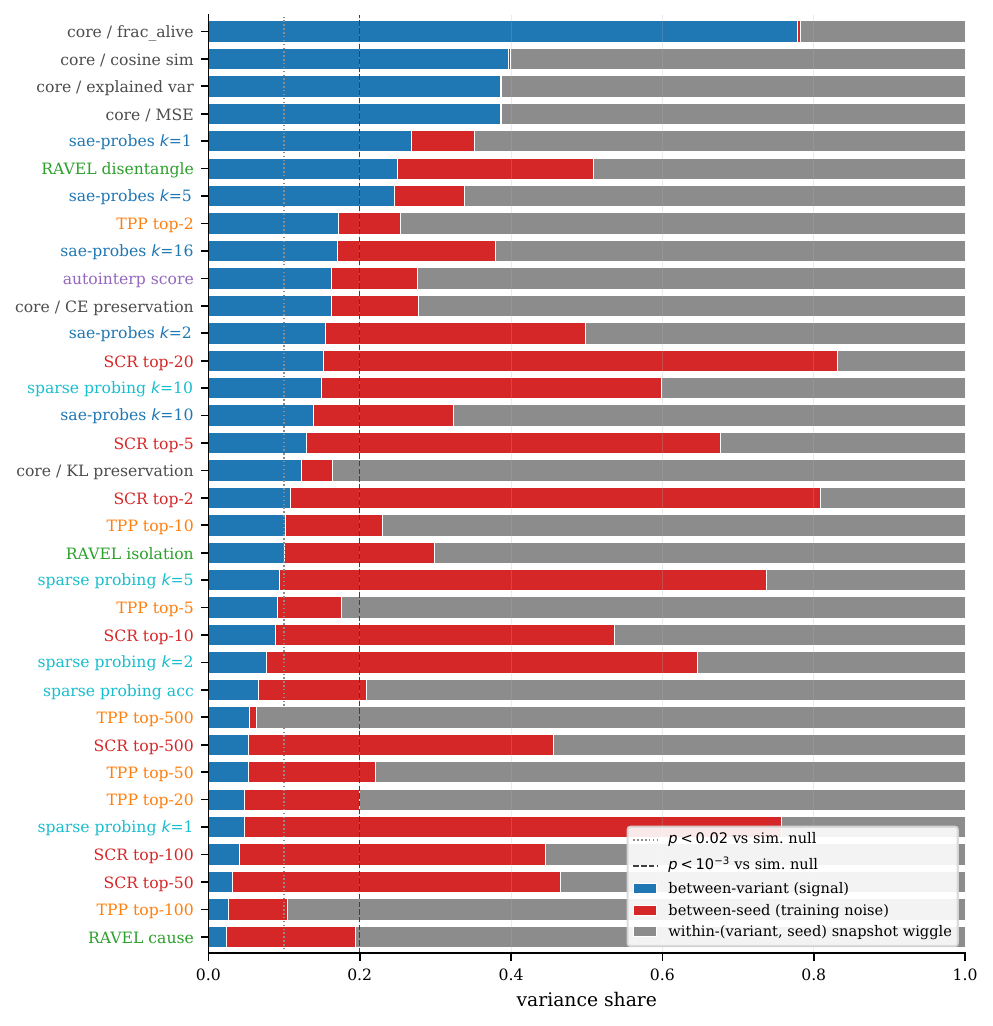}
\caption{Per-metric variance decomposition on the four-variant Matryoshka panel (3 seeds $\times$ 9 post-warmup snapshots per (variant, seed)), sorted by between-variant variance share. The four core reconstruction metrics, sae-probes $k\!\in\!\{1, 5\}$, and RAVEL disentangle are the only metrics where the SAE variant accounts for more than $0.20$ of total variance (dashed line: $p < 10^{-3}$ vs the simulated null; dotted line: $p < 0.02$). Y-axis labels are coloured by metric family (gray = core, blue = sae-probes, cyan = sparse probing, red = SCR, orange = TPP, purple = autointerp, green = RAVEL).}
\label{fig:multi_seed_vd}
\end{figure}

\paragraph{Results.} Table~\ref{tab:multi_seed_vd} reports the variance-share decomposition for the 34 audited metrics, sorted by $\mathrm{share}_\text{variant}$. Figure~\ref{fig:multi_seed_vd} shows the same data as a stacked bar. Seven metrics clear the $0.20$ bar: the four core reconstruction metrics (shares $0.39$--$0.78$), sae-probes at $k\in\{1, 5\}$, and RAVEL disentangle. Twelve more metrics fall in $[0.10, 0.20]$. The remaining fifteen metrics fall below $0.10$, including the three SCR thresholds at $\geq 50$, all four TPP thresholds at $\geq 20$, sparse probing top-$\{1, 5\}$, sparse probing overall, RAVEL cause, and TPP top-$5$. The canonical SCR top-$10$ and TPP top-$10$ thresholds sit at the boundary ($0.09$ and $0.10$ respectively), with most of their score variance attributable to seed and snapshot noise rather than to the SAE variant.

\begin{table}[h]
\caption{Variance-share decomposition for each of the 34 audited SAEBench metrics on the Matryoshka panel, sorted by between-variant share (signal). The right-hand columns are the share of total variance attributable to between-training-seed-within-variant noise and within-(variant, seed) between-snapshot wiggle. The horizontal rules separate metrics with reliable inter-variant signal ($\mathrm{share}_\text{variant} > 0.20$, $p < 10^{-3}$ vs the simulated null), weak evidence ($[0.10, 0.20]$, $p < 0.02$), and no detectable signal ($< 0.10$).}
\label{tab:multi_seed_vd}
\centering
\small
\setlength{\tabcolsep}{6pt}
\begin{tabular}{lrrr}
\toprule
Metric & $\mathrm{share}_\text{variant}$ & $\mathrm{share}_\text{seed}$ & $\mathrm{share}_\text{snap}$ \\
\midrule
core frac\_alive                       & 0.78 & 0.00 & 0.22 \\
core cossim                            & 0.40 & 0.00 & 0.60 \\
core explained\_variance               & 0.39 & 0.00 & 0.61 \\
core MSE                               & 0.39 & 0.00 & 0.61 \\
sae-probes $k$=1                       & 0.27 & 0.08 & 0.65 \\
RAVEL disentangle                      & 0.25 & 0.26 & 0.49 \\
sae-probes $k$=5                       & 0.25 & 0.09 & 0.66 \\
\arrayrulecolor{black!35}\midrule[0.3pt]\arrayrulecolor{black}
TPP top-2                              & 0.17 & 0.08 & 0.75 \\
sae-probes $k$=16                      & 0.17 & 0.21 & 0.62 \\
autointerp                             & 0.16 & 0.11 & 0.72 \\
core CE loss                           & 0.16 & 0.12 & 0.72 \\
sae-probes $k$=2                       & 0.15 & 0.34 & 0.50 \\
SCR top-20                             & 0.15 & 0.68 & 0.17 \\
sparse probing top-10                  & 0.15 & 0.45 & 0.40 \\
sae-probes $k$=10                      & 0.14 & 0.18 & 0.68 \\
SCR top-5                              & 0.13 & 0.55 & 0.32 \\
core KL divergence                     & 0.12 & 0.04 & 0.84 \\
SCR top-2                              & 0.11 & 0.70 & 0.19 \\
TPP top-10                             & 0.10 & 0.13 & 0.77 \\
\arrayrulecolor{black!35}\midrule[0.3pt]\arrayrulecolor{black}
RAVEL isolation                        & 0.10 & 0.20 & 0.70 \\
sparse probing top-5                   & 0.09 & 0.64 & 0.26 \\
TPP top-5                              & 0.09 & 0.08 & 0.82 \\
SCR top-10                             & 0.09 & 0.45 & 0.46 \\
sparse probing top-2                   & 0.08 & 0.57 & 0.35 \\
sparse probing overall                 & 0.07 & 0.14 & 0.79 \\
TPP top-500                            & 0.05 & 0.01 & 0.94 \\
SCR top-500                            & 0.05 & 0.40 & 0.54 \\
TPP top-50                             & 0.05 & 0.17 & 0.78 \\
TPP top-20                             & 0.05 & 0.15 & 0.80 \\
sparse probing top-1                   & 0.05 & 0.71 & 0.24 \\
SCR top-100                            & 0.04 & 0.40 & 0.55 \\
SCR top-50                             & 0.03 & 0.43 & 0.53 \\
TPP top-100                            & 0.03 & 0.08 & 0.89 \\
RAVEL cause                            & 0.02 & 0.17 & 0.81 \\
\bottomrule
\end{tabular}
\end{table}

\subsection{Per-seed winner stability}
\label{app:multi_seed:winners}

For each metric we computed the post-warmup mean per (variant, seed) and picked the variant with the best score (per the metric's natural ``higher = better'' or ``lower = better'' orientation) within each seed. Table~\ref{tab:multi_seed_winners} summarises how many distinct variants are crowned across the three seeds.

\paragraph{Three-distinct winners.} Nine audited metrics produce a different best variant on each of the three seeds: ravel cause, sae-probes $k\!=\!16$, SCR top-$5$, SCR top-$10$, SCR top-$50$, sparse probing top-$1$, TPP top-$10$, TPP top-$20$, and TPP top-$100$. The canonical-batch SCR top-$10$ and TPP top-$\{10, 20\}$ are in this group, so a single-seed report at these canonical thresholds amounts to picking among three plausible winners. This count itself is not statistically beyond a uniform-random-winner null (Pr$(\geq 3$ distinct $) \approx 0.38$ for $4$ variants and $3$ seeds), so we read it as a concrete illustration of the low-$\mathrm{share}_\text{variant}$ result, not as independent evidence.

\paragraph{Single-winner metrics.} Nine audited metrics agree on a single best variant across all three seeds: the four core reconstruction metrics (all pick $n\!=\!1$), core KL divergence ($n\!=\!1$), RAVEL isolation ($n\!=\!1$), TPP top-$500$ ($n\!=\!1$, where the score declines through training so ``best'' is the least-degraded variant), sparse probing overall ($n\!=\!1$, the metric is independent of top-$k$ selection), and sae-probes $k\!=\!5$ ($n\!=\!2$). The remaining sixteen audited metrics pick two distinct winners across the three seeds.

\begin{table}[h]
\caption{Number of distinct variants crowned across 3 training seeds for the 34 audited SAEBench metrics on the Matryoshka panel, with the canonical headline-suite metrics named individually. The ``three distinct'' column lists every metric in that bucket; the ``two distinct'' and ``one consistent'' columns name only canonical or otherwise headline metrics.}
\label{tab:multi_seed_winners}
\centering
\small
\begin{tabular}{lrp{8.2cm}}
\toprule
\# distinct winners across 3 seeds & \# metrics & metrics (selection) \\
\midrule
1 (consistent across all seeds)    & 9          & core MSE, cossim, EV, frac\_alive, KL, CE; RAVEL isolation; sae-probes $k\!=\!5$; TPP top-$500$ \\[0.4em]
2 (different winner from one seed) & 16         & autointerp; RAVEL disentangle; SCR top-$\{2, 20, 100, 500\}$; TPP top-$\{2, 5, 50\}$; sae-probes $k\!\in\!\{1, 2, 10\}$; sparse probing top-$\{2, 5, 10\}$ \\[0.4em]
3 (different winner each seed)     & 9          & ravel cause; sae-probes $k\!=\!16$; SCR top-$\{5, 10, 50\}$; sparse probing top-$1$; TPP top-$\{10, 20, 100\}$ \\
\bottomrule
\end{tabular}
\end{table}

\subsection{Cross-seed ranking agreement}
\label{app:multi_seed:crossseed}

For each metric and each pair of seeds we computed the Spearman $\rho$ between the two seeds' rankings of the four variants by post-warmup mean, then averaged across the three pairs. Table~\ref{tab:multi_seed_rho} reports per-metric mean $\rho$, sorted descending.

The four core reconstruction metrics all reach $\rho = 1.0$ (perfect cross-seed agreement on the variant order $n\!=\!1 > n\!=\!2 > n\!=\!3 > n\!=\!4$, in MSE-direction). Core KL and TPP top-$500$ reach $\rho = 0.87$, also a strong rank-stability signal. Below $\rho \approx 0.5$ the picture decays into the regime where small perturbations to the per-variant means flip the ranking, and below $\rho \approx 0$ the cross-seed ranking is no better than chance.

The lower end of Table~\ref{tab:multi_seed_rho} shows several metrics with negative mean $\rho$, including SCR canonical (top-$10$, $\rho = -0.20$) and TPP canonical (top-$10$ $\rho = -0.20$, top-$20$ $\rho = -0.33$). With only 4 variants, only 11 distinct $\rho$ values are possible per pair, and a uniform-random-rankings null has Pr$($mean $\rho < 0) \approx 0.55$ and Pr$($mean $\rho < -0.33) \approx 0.27$ across 3 seed pairs. We therefore do not interpret negative mean $\rho$ as anti-correlation; the informative reading is that these metrics produce no detectable consistent ranking across seeds, while the metrics at the top of the table do produce a stable ranking.

\begin{table}[h]
\caption{Mean Spearman $\rho$ between the variant rankings produced by each pair of training seeds, averaged across the three seed pairs, for each of the 34 audited SAEBench metrics on the Matryoshka panel. With 4 variants the Spearman lattice has 11 points per pair and 27 distinct mean values, so small differences in mean $\rho$ are quantised. We read the upper end (stable cross-seed ranking) as informative and the lower end (no detectable consistent ranking) as consistent with low $\mathrm{share}_\text{variant}$, not as active anti-correlation.}
\label{tab:multi_seed_rho}
\centering
\small
\setlength{\tabcolsep}{6pt}
\begin{tabular}{lr}
\toprule
Metric & mean $\rho$ across seeds \\
\midrule
core MSE                                    & $+1.00$ \\
core cossim                                 & $+1.00$ \\
core explained\_variance                    & $+1.00$ \\
core frac\_alive                            & $+1.00$ \\
core KL divergence                          & $+0.87$ \\
TPP top-500                                 & $+0.87$ \\
core CE loss                                & $+0.73$ \\
TPP top-2                                   & $+0.73$ \\
autointerp                                  & $+0.67$ \\
RAVEL isolation                             & $+0.67$ \\
sae-probes $k$=1                            & $+0.67$ \\
sae-probes $k$=5                            & $+0.67$ \\
sparse probing overall                      & $+0.47$ \\
sae-probes $k$=16                           & $+0.40$ \\
RAVEL disentangle                           & $+0.20$ \\
TPP top-5                                   & $+0.20$ \\
sparse probing top-10                       & $+0.13$ \\
sparse probing top-5                        & $+0.07$ \\
sae-probes $k$=10                           & $+0.07$ \\
\arrayrulecolor{black!35}\midrule[0.3pt]\arrayrulecolor{black}
TPP top-100                                 & $-0.07$ \\
sae-probes $k$=2                            & $-0.07$ \\
TPP top-50                                  & $-0.07$ \\
sparse probing top-2                        & $-0.13$ \\
SCR top-100                                 & $-0.13$ \\
SCR top-2                                   & $-0.20$ \\
SCR top-5                                   & $-0.20$ \\
RAVEL cause                                 & $-0.20$ \\
SCR top-10                                  & $-0.20$ \\
TPP top-10                                  & $-0.20$ \\
sparse probing top-1                        & $-0.33$ \\
SCR top-500                                 & $-0.33$ \\
TPP top-20                                  & $-0.33$ \\
SCR top-20                                  & $-0.33$ \\
SCR top-50                                  & $-0.40$ \\
\bottomrule
\end{tabular}
\end{table}

\subsection{Caveats}
\label{app:multi_seed:caveats}

Three seeds is the minimum that supports a variance decomposition; the between-seed-within-variant variance is itself estimated from only 3 within-variant samples per metric, so individual share values are imprecise. The qualitative ordering (which metrics clear $\mathrm{share}_\text{variant} > 0.20$) is robust under bootstrap; specific share values should not be over-interpreted, especially in the $[0.05, 0.20]$ band.

The within-trajectory variance term $V_\text{snap}$ includes any monotone trajectory motion within the post-warmup window, so for metrics that improve through training (e.g.\ core MSE) some of $V_\text{snap}$ is real quality drift rather than jitter, biasing $\mathrm{share}_\text{variant}$ down for those metrics. Even with this bias, all four core reconstruction metrics and three encoder-only metrics (sae-probes $k\!\in\!\{1,5\}$, RAVEL disentangle) clear the $0.20$ bar, so the qualitative ordering at the top of the table is robust.

We did not collect multi-seed data for the cross-architecture panel, since we train for so much longer (1.5B tokens vs 300M tokens) on that panel making running multiple seeds too expensive. That panel's variants differ much more than the Matryoshka panel's, so per-variant signal is large enough that a single-seed analysis broadly suffices, but a multi-seed cross-architecture replication remains useful future work.

\section{Full-metric training trajectories}
\label{app:full_traces}

\begin{figure}[h]
\centering
\includegraphics[width=\linewidth]{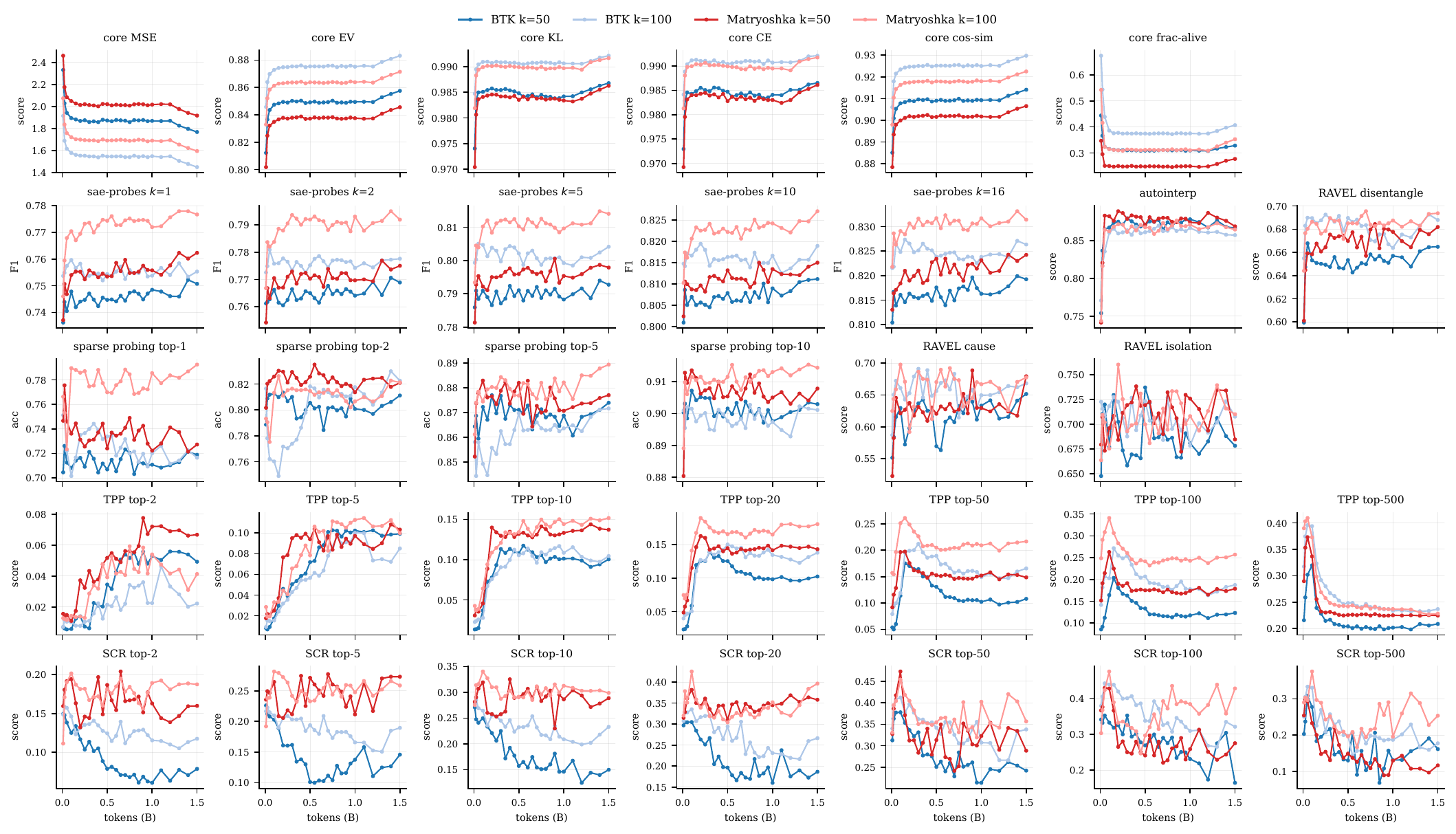}
\caption{Trajectories of every SAEBench metric across the four-SAE architecture training panel (BTK $k\in\{50,100\}$, Matryoshka $k\in\{50,100\}$, 1.5B tokens). Rows from top to bottom: core reconstruction; encoder-only quality (sae-probes, autointerp, RAVEL disentangle); default sparse probing and remaining RAVEL metrics; TPP top-$\{2,5,10,20,50,100,500\}$; SCR top-$\{2,5,10,20,50,100,500\}$.}
\label{fig:arch_traces_full}
\end{figure}

\begin{figure}[h]
\centering
\includegraphics[width=\linewidth]{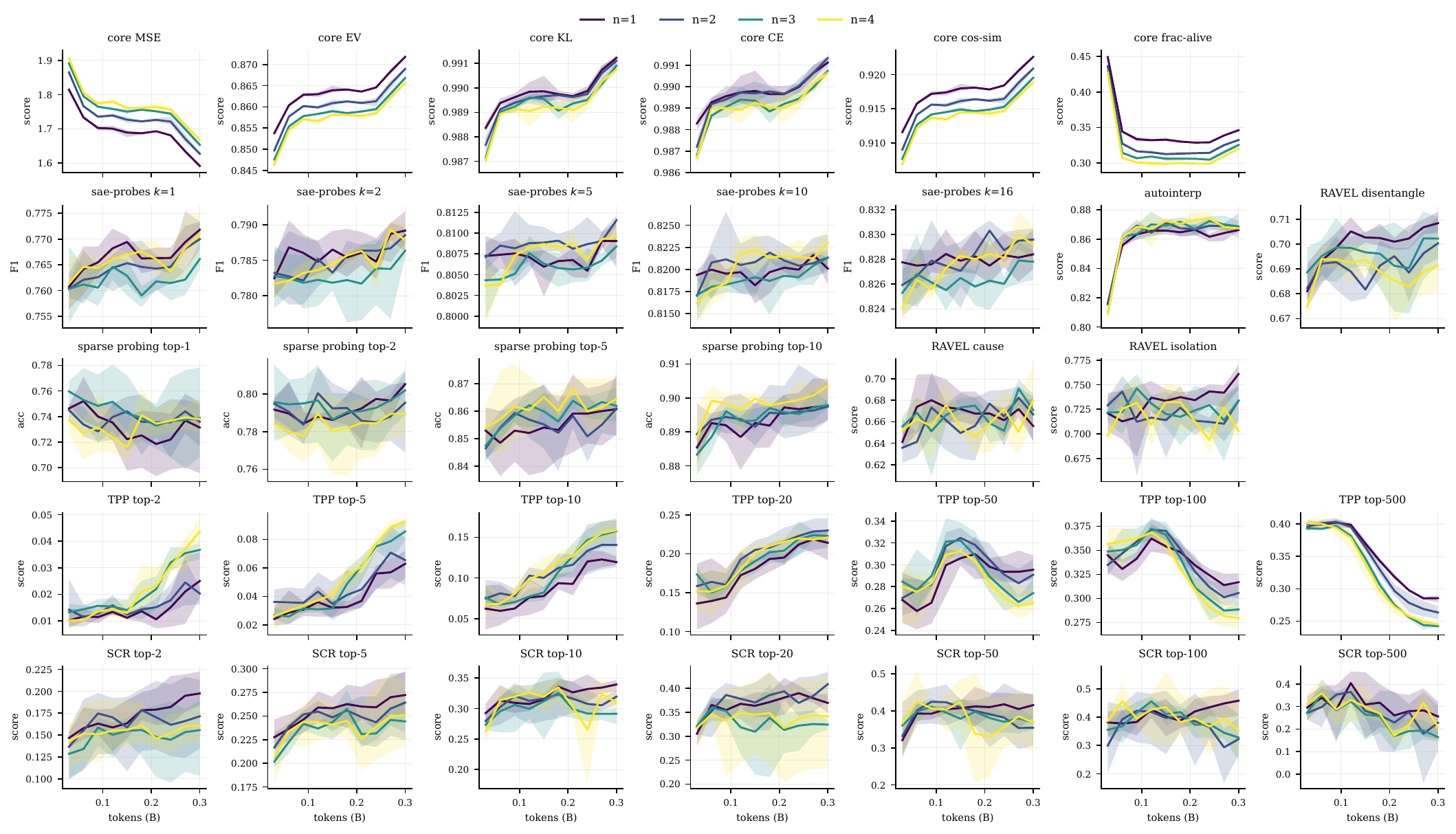}
\caption{Trajectories of every SAEBench metric across the four-SAE sampled-Matryoshka training panel ($n\in\{1,2,3,4\}$ inner widths sampled per training step, 300M tokens, 3 training seeds per variant). Solid lines are seed means; shaded bands are min/max envelopes across the 3 seeds. Layout identical to Figure~\ref{fig:arch_traces_full}.}
\label{fig:sampled_mat_traces_full}
\end{figure}

Figure~\ref{fig:training_traces} in the main text plots four representative metrics across training. For completeness, this appendix gives the trajectories of every SAEBench metric across the same training runs, both for the four-SAE architecture sweep (BatchTopK $k\in\{50,100\}$, Matryoshka $k\in\{50,100\}$; Figure~\ref{fig:arch_traces_full}) and for the four-SAE sampled-Matryoshka hyperparameter sweep ($n \in \{1,2,3,4\}$, varying the number of inner widths sampled per training step; Figure~\ref{fig:sampled_mat_traces_full}). Both panels share the same metric layout: core reconstruction (row 1), encoder-only quality (sae-probes, autointerp, and RAVEL disentangle on row 2), default sparse probing and the remaining RAVEL metrics (row 3), TPP across all thresholds (row 4), and SCR across all thresholds (row 5).

The four-SAE arch panel (Figure~\ref{fig:arch_traces_full}) shows that core, sae-probes, autointerp, RAVEL disentangle, and default sparse probing all rise monotonically through training for every variant; TPP improves at low thresholds (top-$N \leq 20$) and declines at high thresholds (top-$N \geq 50$) for every variant, with the cross-over visible by inspection; SCR declines at most thresholds for both BTK variants and at $\geq 50$ for the Matryoshka variants. RAVEL cause and isolation are nearly flat across training. The four-SAE sampled-Matryoshka panel (Figure~\ref{fig:sampled_mat_traces_full}) compresses to 300M tokens and is noisier, but the same TPP cross-over and SCR-at-large-$N$ decline reproduce.

\section{GT-F1 correlations}
\label{app:f1}

In \S\ref{sec:gt} we mainly focus on correlations against GT-MCC because most metrics seem better correlated to GT-MCC rather than GT-F1, and we only consider it necessary for a proxy metric to correlate to either GT-MCC or GT-F1. Table~\ref{tab:f1} reports the analogous correlations against GT-F1 for the headline benchmarks across the v1 panel; both full-panel and within-trained-SAE values are shown. Finding a proxy metric that correlates with GT-F1 would be a good direction for future SAE benchmarking work.

\begin{table}[h]
\caption{Spearman $\rho$ of each benchmark with GT-MCC and GT-F1 on the v1 panel, multi-seed mean across 3 task-generation seeds at canonical hyperparameters.}
\label{tab:f1}
\centering
\small
\setlength{\tabcolsep}{4pt}
\begin{tabular}{lcc}
\toprule
Benchmark & $\rho$ vs GT-MCC & $\rho$ vs GT-F1 \\
 & (full / within-trained) & (full / within-trained) \\
\midrule
Sparse probing single in-sae (top-16) & $+0.58 / +0.49$ & $+0.02 / -0.22$ \\
Sparse probing boolean in-sae (top-16) & $+0.90 / +0.87$ & $+0.60 / +0.51$ \\
Sparse probing single out-of-sae (top-16) & $+0.23 / +0.19$ & $-0.17 / -0.37$ \\
Sparse probing boolean out-of-sae (top-16) & $-0.04 / -0.22$ & $-0.21 / -0.53$ \\
TPP all in-sae (top-10) & $+0.18 / -0.03$ & $-0.22 / -0.69$ \\
TPP all out-of-sae (top-10) & $+0.38 / +0.42$ & $+0.11 / -0.03$ \\
SCR canonical (top-10) & $+0.64 / +0.65$ & $+0.51 / +0.41$ \\
\bottomrule
\end{tabular}

\end{table}

% \clearpage  % flush pending floats so figures from prior appendices don't drift into the checklist
% \input{checklist.tex}

\end{document}